\documentclass{article} % For LaTeX2e
\usepackage{collas2022_conference,times}

\usepackage{multirow}
\usepackage{amsmath}
\usepackage{amssymb}
\usepackage{mathtools}
\usepackage{amsthm}
\usepackage{booktabs}
\usepackage{algorithm}
\usepackage[noend]{algpseudocode}
\usepackage{wrapfig,lipsum}
\usepackage{enumerate}

\usepackage{amsmath,amsfonts,bm}

\def\eqref#1{equation~\ref{#1}}
\def\1{\bm{1}}

\DeclareMathAlphabet{\mathsfit}{\encodingdefault}{\sfdefault}{m}{sl}
\SetMathAlphabet{\mathsfit}{bold}{\encodingdefault}{\sfdefault}{bx}{n}

\newcommand{\pluseq}{\mathrel{+}=}

\usepackage{hyperref}
\hypersetup{
    colorlinks=true,
    linkcolor=red,
    filecolor=magenta,      
    urlcolor=blue,
    citecolor=purple,
    pdftitle={Overleaf Example},
    pdfpagemode=FullScreen,
    }

\makeatletter
\newcommand{\printfnsymbol}[1]{%
  \textsuperscript{\@fnsymbol{#1}}%
}

\title{Consistency is the key to further mitigating catastrophic forgetting in continual learning}

\author{Prashant Bhat, Bahram Zonooz\thanks{Shared last author.}, Elahe Arani\printfnsymbol{1} \\
Advanced Research Lab, NavInfo Europe, Eindhoven, The Netherlands\\
\texttt{\{prashant.bhat, elahe.arani\}@navinfo.eu, bahram.zonooz@gmail.com}
}

\usepackage[capitalize]{cleveref}
\crefname{section}{Sec.}{Secs.}
\Crefname{section}{Section}{Sections}
\Crefname{table}{Table}{Tables}
\crefname{table}{Tab.}{Tabs.}

\collasfinalcopy % Uncomment for camera-ready version, but NOT for submission.

\begin{document}

\maketitle

%%%%%%%%% ABSTRACT
\begin{abstract}
   Deep neural networks struggle to continually learn multiple sequential tasks due to catastrophic forgetting of previously learned tasks. Rehearsal-based methods which explicitly store previous task samples in the buffer and interleave them with the current task samples have proven to be the most effective in mitigating forgetting. However, Experience Replay (ER) does not perform well under low-buffer regimes and longer task sequences as its performance is commensurate with the buffer size. Consistency in predictions of soft-targets can assist ER in preserving information pertaining to previous tasks better as soft-targets capture the rich similarity structure of the data. Therefore, we examine the role of consistency regularization in ER framework under various continual learning scenarios. We also propose to cast consistency regularization as a self-supervised pretext task thereby enabling the use of a wide variety of self-supervised learning methods as regularizers. While simultaneously enhancing model calibration and robustness to natural corruptions, regularizing consistency in predictions results in lesser forgetting across all continual learning scenarios. Among the different families of regularizers, we find that stricter consistency constraints preserve previous task information in ER better. 
  \footnote{Code can be found at:  \url{https://github.com/NeurAI-Lab/ConsistencyCL}.}

\end{abstract}

%%%%%%%%% BODY TEXT
\section{Introduction}

% Learning tasks sequentially by accumulating knowledge learned in the past is one of the hallmarks of human intelligence.
Continual Learning (CL) refers to a learning paradigm where computational systems sequentially learn multiple tasks with data becoming progressively available over time. An ideal CL system must be plastic enough to integrate novel information and stable enough to not interfere with the consolidated knowledge \citep{parisi2019continual}. In deep neural networks, however, sufficient plasticity to acquire new tasks results in large weight changes disrupting consolidated knowledge, referred to as catastrophic forgetting \citep{goodfellow2013empirical, mccloskey1989catastrophic}. Catastrophic forgetting often leads to a swift drop in performance and in the worst case, the previously learned information is completely overwritten by the new one \citep{parisi2019continual}. The menace of catastrophic forgetting is not limited to CL alone, but manifests in multitask learning \citep{kudugunta2019investigating}, and supervised learning under domain shift \citep{ovadia2019can} as well.

Several approaches have been proposed in the literature to address the stability-plasticity dilemma, the root cause of catastrophic forgetting in CL in neural networks. 
% Inspired by the theoretical neuroscience findings that forgetting can be mitigated through synapses with a cascade of states yielding different levels of plasticity \citep{fusi2005cascade},  
Weight-regularization methods \citep{zenke2017continual, schwarz2018progress} impose explicit constraints on the neural network updates via an additional regularization term thereby restricting the change in weights pertaining to previous tasks. Although fairly successful in some CL scenarios, these methods fail in scenarios such as class-incremental learning. Parameter-isolation methods \citep{rusu2016progressive} allocate a distinct set of parameters to distinct tasks to minimize the interference. However, these methods do not scale well with a large number of tasks. Rehearsal-based methods \citep{gem,agem} explicitly store and replay a subset of previous task samples along  with the current batch of samples, have proven to be most effective in minimizing interference in challenging CL tasks \citep{farquhar2018towards}. 

Simple Experience Replay (ER) \citep{ratcliff1990connectionist, robins1995catastrophic} which interleaves old task samples with the current task samples is one of the top-performing methods across different CL scenarios \citep{buzzega2020dark}. There might be scenarios where large memory footprint is not as problematic, in this work we focus on a specific problem setting where large memory footprint is expensive. Given such a scenario, ER does not perform well under low-buffer regimes and longer task sequences as its performance is commensurate with the buffer size.  To preserve the information pertaining to previous tasks better, soft-targets can be leveraged as they contain more information than the hard targets and capture the rich similarity structure of the data \citep{hinton2015}. 
% To further leverage information stored in the buffer, 
% To preserve the information pertaining to previous tasks better, 
Therefore, several works \citep{fdr, buzzega2020dark, arani2022learning, pham2021dualnet} leverage soft-targets in addition to hard ones thereby drawing more information from the limited buffered samples. Despite differences in their architecture and training paradigm, these methods have one thing in common: enforcing consistency in predictions across time-separated views of buffered samples across tasks. By regularizing consistency in predictions, these approaches preserve rich information about the previous tasks leading to a further reduction in forgetting. 

\begin{figure}[t]
\center
  \includegraphics[width=1\textwidth]{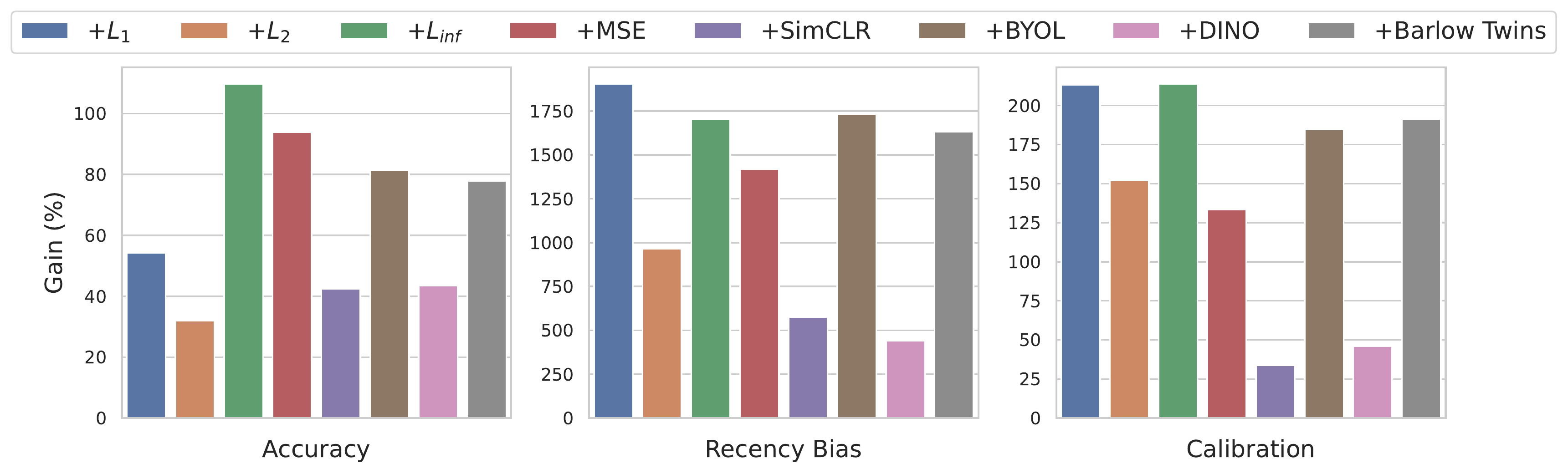}
\caption{Relative performance gain of different consistency regularizers added to ER over vanilla ER. All models are trained on S-TinyImageNet with buffer size 500. Higher value indicates better performance gain. More details can be found in the Appendix \ref{appendix_overview}.}
\label{fig:overview}
\end{figure}

 Consistency regularization has been widely used in semi-supervised learning to ensure that model's output is unaffected for samples augmented in a semantic-preserving way \citep{xie2020unsupervised, zhai2019s4l, sajjadi2016regularization}.
% it is sparingly used in the CL literature. More importantly, 
However, a comprehensive study to understand the role of consistency regularization in CL under a common framework does not exist. Therefore, we examine the role of consistency regularization in CL under a common framework: we store the model's predictions along with the input image and corresponding ground-truth in the buffer. In the subsequent training iterations, we penalize the CL model for its deviation in predictions. We explore Minkowski distance functions ($L_p$), Kullback-Leibler Divergence (KL-Div), and Mean-Squared Error (MSE) as consistency regularizers. In addition, we cast regularizing consistency as a pretext task of bringing closer the corresponding predictions in the representational space by employing state-of-the-art Self-Supervised Learning (SSL) methods. This gives us the design flexibility to consider a wide variety of methods as regularizers thereby expanding the ambit of our analyses.

We find that any form of consistency regularization is the key to further leveraging buffered information in ER across all different CL scenarios.  Moreover, stricter consistency constraints preserve the previous task information better enabling the CL model to forget less. Our empirical results show that Minkowski distance functions, extremely simple and intuitive are surprisingly one of the most effective families in alleviating catastrophic forgetting. Additionally, regularizing consistency in predictions leads to a more robust and well-calibrated model. Consistency regularization also mitigates the task recency bias thereby leading to more evenly distributed task prediction probabilities. Figure \ref{fig:overview} provides a brief overview of some of our results.

% Our contributions are as follows:
% \begin{itemize}
%     \item Under ER framework, we examine the role of regularizing consistency in predictions in CL. We compare the performance of different family of regularizers under various CL scenarios.
%     \item By courting consistency regularization as a self-supervised pretext task, we explore recent state-of-the-art SSL algorithms as regularizers in CL. This gives the opportunity to investigate wide variety of functions expanding the ambit of our analyses. 
%     \item We also provide an in-depth model analysis in terms of model calibration, task recency bias, and robustness to natural corruptions. 
% \end{itemize}

\section{Related works}
CL has remained one of the long-standing challenges for neural networks due to catastrophic forgetting \citep{HASSABIS2017245}. The stability-plasticity dilemma, the root cause of catastrophic forgetting, has been well studied in both biological and artificial neural networks \citep{Bugaiska, Grossberg1982}. Early works attempted to minimize forgetting in neural networks through experience rehearsal \citep{ratcliff1990connectionist, robins1995catastrophic, gem, agem} by interleaving samples belonging to previous tasks in the replay buffer with the current batch of samples while training on the new tasks. Simple ER \citep{ratcliff1990connectionist, robins1995catastrophic} is one of the top-performing methods across different CL scenarios \citep{buzzega2020dark}. In this work we focus on a specific problem setting where storing large buffers can be expensive. Given such a scenario, ER fails to perform well under low-buffer regimes and longer task sequences.
% There can be scenarios where storing large buffers can be expensive. 
% However, as the buffer size reduces, the performance of the ER also drop drastically. 

To preserve information pertaining to previous tasks better, several other methods build on top of ER: Function Distance Regularization (FDR) \citep{fdr} used mean-squared error to align current outputs with the past exemplars. Dark-experience Replay (DER) \citep{buzzega2020dark} followed the suit and extended the regularization to different CL scenarios. Inspired by the complementary-learning system (CLS) theory \citep{kumaran2016learning} in the brain, several methods include fast and slow learning networks to consolidate the knowledge better: CLS-ER \citep{arani2022learning} used  dual memory experience replay to maintain short-term and long-term semantic memories. Despite differences in their approach and underlying architecture, consistency regularization unifies them under one hood: these approaches regularize consistency in predictions of buffered samples across tasks thereby preserving rich information about the previous tasks. 

As the training progresses, soft-targets carry more information per training sample than the hard targets. Although this has little influence on the cross-entropy objective, it captures the rich similarity structure of the data \citep{hinton2015}. Therefore, consistency regularization on soft-targets has been a widely used technique in many domains including semi-supervised learning, self-supervised learning, and knowledge distillation \citep{bachman2014learning,zhai2019s4l,chen2020simple,hinton2015}.  In knowledge distillation, several approaches \citep{lwf, icarl} employ past version of the CL model as a teacher and distill knowledge to the current model. Learning without forgetting (LwF) \citep{lwf} reduces the feature drift by smoothening the predictions at the beginning of each task without employing the memory buffer. In semi-supervised learning, the core idea is simple: input image is perturbed in semantic-preserving ways and the model's sensitivity to perturbations is penalized. Consistency regularizer forces the model to learn representations invariant to semantic-preserving perturbations. These perturbations can manifest in many ways: they can be augmentations such as random cropping, Gaussian noise, colorization, or even adversarial attacks. The regularization term generally in semi-supervised learning is either mean-squared error \citep{sajjadi2016regularization} between model's output of perturbed and non-perturbed images or KL-divergence \citep{miyato2018virtual} between distribution over classes implied by the logits. 
However, it is difficult to gauge the role of consistency regularization with a limited number of regularizers. On the other hand, SSL objectives \citep{chen2020simple, he2020momentum} implicitly impose a loose form of consistency: they enforce the representations of the multiple augmented views to be close by, not necessarily restricting them to be the same in the latent space. Although they have been used for pre-training, these methods have not been used as explicit consistency regularizers. Moreover, a diverse family of SSL functions provides design flexibility to better understand the role of consistency regularization.

Consistency regularization is relatively new to CL and has not been well explored. Therefore, we study the effectiveness of consistency regularization in CL and investigate its role in mitigating catastrophic forgetting under a common framework.

\begin{figure}[!t]
\begin{minipage}{.55\textwidth}
\begin{algorithm}[H]
\caption{ER with consistency regularization}
\label{alg:method}
\begin{algorithmic}[1]
    \Statex \textbf{input: } Data streams $\mathcal{D}_{t} \;\; \forall \; t \; \in \{1,...,T $\}, Model $f_\theta$
    \Statex ~~~~~~~~~~~~~Balancing factors $\alpha$ and $\beta$, Memory buffer $\mathcal{D}_m = \{\}$
    
    \ForAll {tasks $t \in \{1, 2,..,T\}$} 
        \For{minibatch $\{x_i, y_i\}_{i=1}^B \in \mathcal{D}_{t}$}
        \State $\mathcal{L}_{er} = 0$
        \State $\hat{y}_i = f_\theta(x_i)$
        \State Compute $\mathcal{L}_{er} \; \pluseq \frac{1}{B} \displaystyle \sum_{B} \mathcal{L}_{ce} (\hat{y}_{i},  y_{i})$
            \If{$\mathcal{D}_{m} \neq \emptyset$ }
                \For{ minibatch $\{x_{j}, y_{j}, z_{j}\}_{j=1}^B \in \mathcal{D}_{m}$}
                    % \State $x_{j}^{'} = augment(x_{j}^{'})$ 
                    \State $\hat{y}_{j} = f_\theta(x_{j})$
                    \State Compute $\mathcal{L}_{er} \; \pluseq \frac{\alpha}{B}  \displaystyle \sum_{B} \mathcal{L}_{ce} (\hat{y}_{j},  y_{j})$
                \EndFor
                \For{ minibatch $\{x_j, y_j, z_j\}_{j=1}^B \in \mathcal{D}_{m}$}
                    % \State $x_j = augment(x_j)$ 
                    \State $\hat{z}_j = f_\theta(x_j)$
                    \State Compute $\mathcal{L}_{cr}$
                \EndFor
            \EndIf
            % \State $x_i = augment(x_i)$ 
            \State Compute $\mathcal{L} = \mathcal{L}_{er} + \beta  \mathcal{L}_{cr}$
            \State Compute the gradients $\displaystyle \frac{\delta \mathcal{L}}{\delta \theta}$
            \State Update the model $f_{\theta}$
            \State Update the memory buffer $\mathcal{D}_m$
        \EndFor
    \EndFor
    \State \Return{model $f_\theta$ }
    \end{algorithmic}
    \end{algorithm}
\end{minipage}% This must go next to `\end{minipage}`
\hspace{0.03\textwidth}%
\begin{minipage}{.42\textwidth}
    \begin{figure}[H]
    \centering
    \includegraphics[width=.92\linewidth]{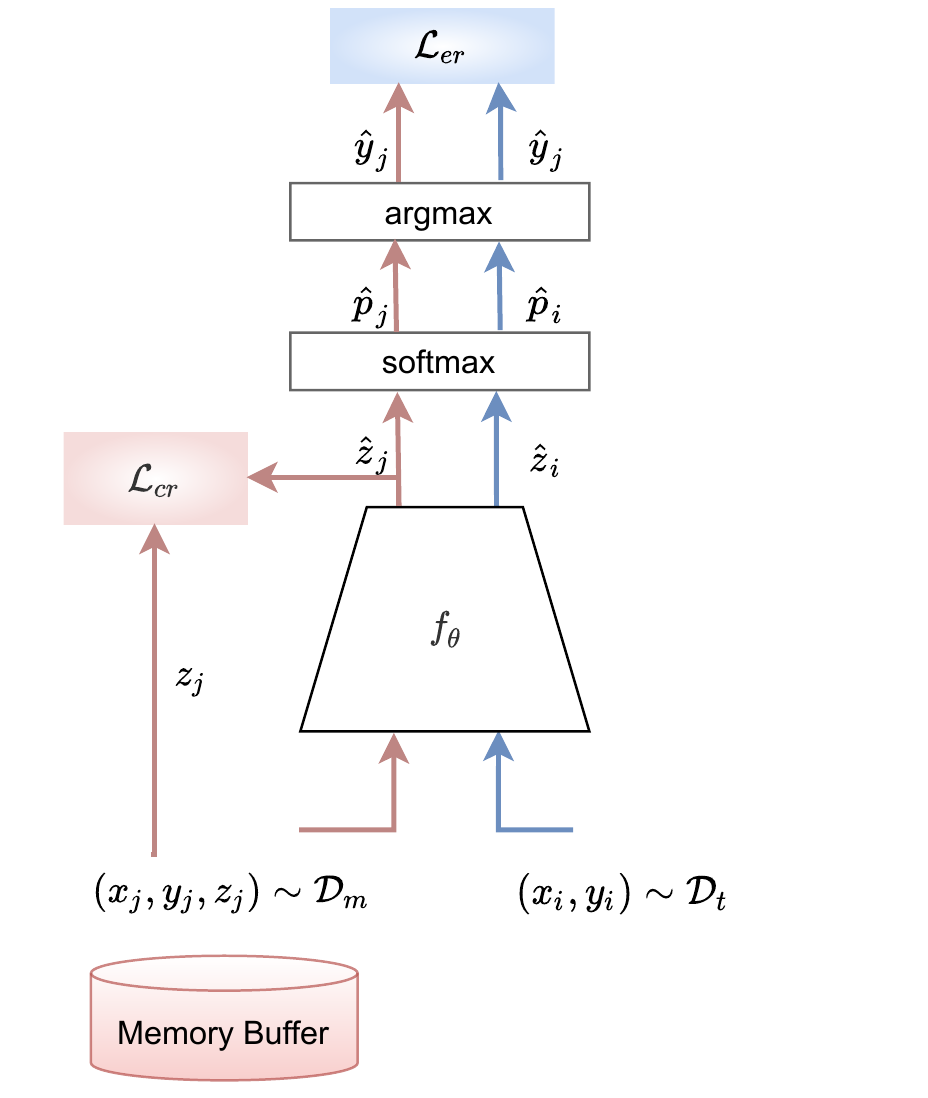}
      \caption{Schematic diagram of the framework. In addition to ER objective ($\mathcal{L}_{er}$), we penalize the model for deviation from its past predictions through consistency regularization ($\mathcal{L}_{cr}$).}
      \label{fig:main}
    \end{figure}
\end{minipage}
\end{figure}

\section{Experience Replay with consistency regularization}
CL normally consists of $T$ sequential tasks indexed by $t \in \{1, 2, . . . , T\}$.  During each task, $N_t$ input samples and the corresponding labels $\{(x_{i}, y_{i})\}_{i=1}^{N_t}$ are drawn from the task data distribution $\mathcal{D}_{t}$. The CL model $f_{\theta}$  consists of a backbone network (such as ResNet-18 \citep{he2016deep}) and a linear classifier representing all classes. The model $f_{\theta}$ is sequentially optimized on data stream of one task with the cross-entropy objective function. CL is especially challenging since the data from the previous tasks are unavailable, i.e at any point during training the model $f_{\theta}$ has access to the current data distribution $\mathcal{D}_t$ alone. As the cross-entropy objective is solely optimized for the current task, plasticity overtakes stability resulting in overfitting on the current task and catastrophic forgetting of older tasks.

ER sought to address this problem by storing a subset of training data from previous tasks (into a memory buffer) and replaying them alongside $\mathcal{D}_{t}$:
\begin{equation}
\label{eqn_er}
    \mathcal{L}_{er} \triangleq  \displaystyle \mathop{\mathbb{E}}_{(x_{i}, y_{i}) \sim \mathcal{D}_{t}} \left[ \: \mathcal{L}_{ce} (\hat{y}_i, y_i) \: \right] + \alpha \displaystyle \mathop{\mathbb{E}}_{(x_j, y_j) \sim D_{m}} \left[ \; \mathcal{L}_{ce} (\hat{y}_j, y_j) \; \right]
\end{equation}
where $\mathcal{L}_{ce}$ is a cross-entropy objective and $\alpha$ is a balancing factor.
$\mathcal{D}_{m}$ represents the distribution of samples stored in the memory buffer. We use reservoir sampling \citep{vitter1985random} to populate the memory buffer in order not to rely on the task boundaries. Reservoir sampling selects random samples with equal probability from the input stream throughout the CL training, is suited for general CL without task boundary information. ER partially improves the stability-plasticity dilemma through twin objectives: supervisory signal from $\mathcal{D}_t$ improves plasticity while that from $\mathcal{D}_m$ ameliorates the stability, thus partially addressing catastrophic forgetting. In practice, only a limited number of samples are stored in the memory buffer owing to budget constraints ($|D_{t}| \gg |D_{m}|$). As the performance of ER is commensurate with the buffer size, it is quintessential to derive as much information as possible about the previous tasks from limited buffer samples. 

Consistency regularization plays a pivotal role in approximating the past behaviour by enforcing consistency across time-separated current and past predictions of buffered samples. Through consistency, rich learned information about the previous tasks can be better reinforced resulting in less forgetting. 
In addition, enforcing consistency between the corresponding pre-softmax responses has a beneficial effect: it avoids information loss occurring due to softmax operation. Therefore, the final learning objective of ER with consistency regularization is defined as follows:
\begin{equation}
\label{eqn_final}
     \mathcal{L} = \mathcal{L}_{er} + \beta \mathcal{L}_{cr}
\end{equation}
where $\beta$ is a balancing factor and $\mathcal{L}_{cr}$ represents any consistency regularizer on the predictions. Figure \ref{fig:main} provides a schematic diagram of the ER framework with consistency regularization. The detailed training regime is in Algorithm \ref{alg:method}. In the next section, we provide details about the various choices for $\mathcal{L}_{cr}$.

\section{Consistency regularizers}
Consistency regularization can be straightforwardly defined with the expected Minkowski distance $(L_p)$ between the corresponding pairs of predictions: 
\begin{equation}
\label{eqn_lp}
    \mathcal{L}_{cr} \triangleq \displaystyle \mathop{\mathbb{E}}_{(x_j, y_j, z_j) \sim D_{m}} \lVert \hat{z_j} - z_j \rVert_{p}
\end{equation}
where $z_j$ represents model's pre-softmax responses stored in the buffer. Squared-$L_2$ distance can also be viewed as Mean-Squared Error (MSE), up to a constant factor. Under mild assumptions \citep{hinton2015},  the optimization of the KL-divergence is equivalent to minimizing the Euclidean distance between the corresponding logits. Therefore, $\mathcal{D}_{KL} (\hat{p_j} \; \lVert \; p_j)$ can also be used in place of $L_p$ in Equation \ref{eqn_lp}.

Enforcing consistency in CL is akin to solving a pretext task of bringing current and past exemplar outputs closer in the representational space. 
% Therefore, we propose to cast consistency regularization as an auxiliary self-supervised pretext task. 
It can be simply achieved through maximizing mutual information $I(\hat{z}_j, z_j)$ as follows: 
\begin{equation}
\label{eqn_mi}
     \mathcal{L}_{cr} \triangleq - \displaystyle \sum_{\hat{z}_j} \sum_{z_j} Q_{\hat{j}j} \ln { \frac{Q_{\hat{j}j}}{Q_{\hat{j}} \; Q_{j} }}
\end{equation}
where the conditional joint distribution can be approximated through $Q_{\hat{j}j} = p(\hat{z}_j, z_j | x_j) = Q(\hat{p}_j) \cdot Q(p_j) $. The marginal distributions $Q_{\hat{j}} = Q(\hat{p}_j)$ and $Q_{j} = Q(p_j)$ can be obtained by summing over rows and columns of $Q_{j\hat{j}}$ matrix. With more tasks and more classes per task, estimating mutual information in high dimensional space becomes a notoriously difficult task, and in practice one often maximizes a tractable lower bound on this quantity \citep{poole2019variational}. InfoNCE \citep{oord2018representation}, based on Noise Contrast Estimation \citep{gutmann2012noise}, is one such lower bound that has been shown to work well in practice: 
\begin{equation}
\label{eqn_infonce}
   I(\hat{z}_j, z_j) \geq \displaystyle \mathop{\mathbb{E}}_{q(\hat{z}_j, z_j)} [ \; <\hat{z}_j, z_j> \;
   - \; \displaystyle \mathop{\mathbb{E}}_{q(Z)} log \sum_{z_k \in Z} exp \; <\hat{z}_j, z_k> \; ] \;  + \; log |Z|
\end{equation}

\begin{wraptable}{r}{0.45\linewidth}
\vspace{-20pt}
\caption{Overview of various consistency regularizers used in this work.}
\label{tab:ssl_algorithms}
\begin{small}
\center
\begin{tabular}{l|l|l}
    \toprule
    \multicolumn{2}{l|}{Category} &  Methods  \\
    \midrule 
    \multicolumn{2}{l|}{Minkowski distance functions} & $L_{1}$, $L_{2}$,  $L_{inf}$  \\
    \multicolumn{2}{l|}{} & MSE = $\frac{1}{N} L^2_2$ \\
    \midrule
    \multicolumn{2}{l|}{Kullback-Leibler Divergence} & $\mathcal{D}_{KL} (\hat{p_j} \; \lVert \; p_j)$   \\
    \midrule
    \multirow{4}{*}{\begin{tabular}[l]{@{}c@{}}SSL methods\end{tabular}} & InfoNCE & SimCLR \\
     & Mean squared error & BYOL \\
     & Cross-Entropy & DINO \\
     & Cross-Correlation & Barlow Twins \\
\bottomrule
\end{tabular}
\end{small}
\end{wraptable}
where $z_k \in Z \setminus \{\hat{z_j}, z_j\}$ is a set of negatives and $<.,.>$ computes cosine similarity. InfoNCE has recently been popular among SSL methods \citep{chen2020simple, he2020momentum}. Courting regularization of consistency in model's predictions as an auxiliary self-supervised pretext task provides the design flexibility as  $\mathcal{L}_{cr}$ can be replaced with any state-of-the art SSL method in our framework. Table \ref{tab:ssl_algorithms} presents the overview of regularizers along with the state-of-the-art SSL method under four broad categories. This lets us examine different families of functions as regularizers expanding the ambit of our forthcoming analyses. Implementation details of these approaches are in Appendix \ref{app_ssl_algo}.

% \section{Results}

\section{Experimental setup}
Following \citep{van2019three}, we evaluate consistency regularization added to ER on class incremental learning (Class-IL), domain incremental learning (Domain-IL), and MNIST-360 \citep{buzzega2020dark} scenarios. In Class-IL, the CL model encounters a new set of classes in each task and learns to distinguish all classes encountered so far after each task. In Domain-IL however, the number of classes remains the same across subsequent tasks, but the input distribution is shifted. MNIST-360 exposes the CL model to both sharp class distribution shift and smooth rotational distribution shift in MNIST-360. More information about the datasets is detailed in Appendix \ref{datasets}. 

We build upon Mammoth \citep{buzzega2020dark} CL repository in PyTorch. We employ ResNet-18 for Class-IL and a 2-layer fully-connected network of 100 neurons each with ReLU activation for Domain-IL and MNIST-360 tasks. To ensure a fair comparison between different regularizers, the training settings have been standardized: backbone, number of epochs training for each task, batch size, reservoir buffer and buffer size have been kept the same for all methods along with the results averaged over three random seeds. We consider $L_1$, $L_2$, and $L_{inf}$ from the Minkowski distance family, MSE, KL-Div, MI, and four different SSL algorithms SimCLR, BYOL, DINO, and Barlow Twins (Table \ref{tab:ssl_algorithms}) as consistency regularizers in addition to ER. Since not all regularizers are used in their original form, we use a shorthand notation (e.g. +SimCLR) to refer to them being used as a regularizer on top of ER. Implementation details of these approaches can be found in Appendix \ref{app_ssl_algo}. In the empirical results, we also provide a lower bound \textit{SGD}, without any help to mitigate catastrophic forgetting, and an upper bound \textit{Joint}, where training is done using entire dataset. 

\begin{table} [t]
\centering
\caption{Comparison of Top-1 Accuracy ($\%$) of vanilla ER and ER plus various consistency regularization under different challenging CL scenarios. Best result is in bold and the second best is underlined.}
\label{tab:main}
% \resizebox{\textwidth}{!}{
\begin{tabular}{ll|ccc|c|c}
\toprule
\multirow{2}{*}{\begin{tabular}[l]{@{}c@{}}Buffer\\ size\end{tabular}} & \multirow{2}{*}{Method}  & \multicolumn{3}{c|}{Class-IL}  & Domain-IL &  \\
\cmidrule{3-7}
&& S-CIFAR-10 & S-CIFAR-100 & S-TinyImageNet & R-MNIST & MNIST-360\\
\midrule
 
\multirow{2}{*}{-}  & Joint  &  92.20 \scriptsize{$\pm$ 0.15}   & 70.56 \scriptsize{$\pm$ 0.28}  & 59.99 \scriptsize{$\pm$ 0.19} & 96.52 \scriptsize{$\pm$ 0.12} & 82.05 \scriptsize{$\pm$ 0.62}  \\

& SGD & 19.62 \scriptsize{$\pm$ 0.05}  & 17.49 \scriptsize{$\pm$ 0.28} &   07.92 \scriptsize{$\pm$ 0.26} & 70.76 \scriptsize{$\pm$ 5.61} & 21.09 \scriptsize{$\pm$ 0.21}\\ \midrule
    
    \multirow{11}{*}{200}  & ER  & 48.19 \scriptsize{$\pm$ 1.37} &  21.40 \scriptsize{$\pm$ 0.22} &  8.57  \scriptsize{$\pm$ 0.04} & 82.09	 \scriptsize{$\pm$ 3.32} & 51.76 \scriptsize{$\pm$ 2.19 }\\
    
    & +$L_{1}$  &  \textbf{65.33} \scriptsize{$\pm$ 1.24} &   	28.74 \scriptsize{$\pm$ 1.92}  & 10.29 \scriptsize{$\pm$ 0.64} & 86.99 \scriptsize{$\pm$ 0.60} & 57.28 \scriptsize{$\pm$ 2.04}\\
    
    & +$L_{2}$  &  63.04 \scriptsize{$\pm$ 1.44} &  29.72 \scriptsize{$\pm$ 1.16}  &   9.32 \scriptsize{$\pm$ 0.38} &   89.72 \scriptsize{$\pm$ 0.19} & \textbf{61.68} \scriptsize{$\pm$ 1.17}\\
    
    & +$L_{inf}$  & \underline{63.77}  \scriptsize{$\pm$ 1.25}  &  \textbf{32.28} \scriptsize{$\pm$ 2.52} &   \textbf{13.09} \scriptsize{$\pm$ 0.47} & \underline{90.01} \scriptsize{$\pm$ 0.13}  & \underline{61.03}  \scriptsize{$\pm$ 1.57} \\
    
    & +KL-Div  &  57.78 \scriptsize{$\pm$ 1.42}  &  21.86 \scriptsize{$\pm$ 0.25} &    8.46 \scriptsize{$\pm$ 0.27} & 84.69 \scriptsize{$\pm$ 0.70} & 50.28 \scriptsize{$\pm$ 0.59}\\
    
    & +MSE  & 65.12  \scriptsize{$\pm$ 1.61}  &  	29.60 \scriptsize{$\pm$ 1.14} &  11.14   \scriptsize{$\pm$ 0.63}  & \textbf{90.78} \scriptsize{$\pm$ 0.22} & 59.04 \scriptsize{$\pm$ 1.69}\\

  & +MI  &  59.56 \scriptsize{$\pm$ 1.35}  & 	22.49 \scriptsize{$\pm$ 0.28} &  8.45 \scriptsize{$\pm$ 0.24} & 86.37 \scriptsize{$\pm$ 0.10} & 50.84 \scriptsize{$\pm$ 1.74}\\

  & +SimCLR  &  	60.19 \scriptsize{$\pm$ 2.54}  &  	24.75 \scriptsize{$\pm$ 0.6}  &   10.16 \scriptsize{$\pm$  0.33 } & 85.79 \scriptsize{$\pm$ 0.99} & 50.66 \scriptsize{$\pm$ 2.23}\\

 & +BYOL  &  63.57 \scriptsize{$\pm$ 0.63}  &  \underline{31.23} \scriptsize{$\pm$ 1.45} &    12.15 \scriptsize{$\pm$ 0.47} & 89.26 \scriptsize{$\pm$ 0.14} & 54.42 \scriptsize{$\pm$ 2.94} \\

 & +DINO  &  63.03 \scriptsize{$\pm$ 2.08}  &  26.21 \scriptsize{$\pm$ 0.55} &   9.87 \scriptsize{$\pm$ 0.43} & 86.78 \scriptsize{$\pm$ 0.72} & 47.16 \scriptsize{$\pm$ 2.36} \\

 & +Barlow Twins  &  62.91 \scriptsize{$\pm$ 1.01}  &  26.51 \scriptsize{$\pm$ 0.78}  & \underline{12.20} \scriptsize{$\pm$ 0.24} & 88.65 \scriptsize{$\pm$ 0.37} & 53.02 \scriptsize{$\pm$ 0.91} \\
 
\midrule  
    
\multirow{11}{*}{500}  & ER  & 60.93 \scriptsize{$\pm$ 1.50} &   28.02 \scriptsize{$\pm$ 0.31}  &   10.08 \scriptsize{$\pm$ 0.21} &  88.38 \scriptsize{$\pm$ 1.54} & 63.78 \scriptsize{$\pm$ 2.70}  \\
  
& +$L_{1}$  &  \textbf{74.39} \scriptsize{$\pm$ 0.54 }  &   40.40\scriptsize{$\pm$ 1.31}  &   15.59 \scriptsize{$\pm$  0.17}  & 89.96  \scriptsize{$\pm$ 0.39} & 68.77 \scriptsize{$\pm$ 1.32} \\

& +$L_{2}$  & 72.95  \scriptsize{$\pm$ 0.74} &    40.92 \scriptsize{$\pm$ 0.95}  &   13.77 \scriptsize{$\pm$ 0.30}  & 91.79 \scriptsize{$\pm$ 0.63} & \textbf{73.32} \scriptsize{$\pm$ 0.30}  \\

& +$L_{inf}$  &  \underline{73.83} \scriptsize{$\pm$ 0.35}  &   \textbf{42.15} \scriptsize{$\pm$ 1.88} &   \textbf{20.47} \scriptsize{$\pm$ 0.99}  & \underline{92.22} \scriptsize{$\pm$ 0.62} & 70.89 \scriptsize{$\pm$ 0.70}  \\

& +KL-Div  &  70.50 \scriptsize{$\pm$ 0.48} &   27.91 \scriptsize{$\pm$ 0.98} &   10.22   \scriptsize{$\pm$ 0.17}  & 88.38 \scriptsize{$\pm$ 1.52} & 65.50 \scriptsize{$\pm$ 2.84}  \\
    
& +MSE  & 73.60  \scriptsize{$\pm$ 0.72} &  \underline{41.40} \scriptsize{$\pm$ 0.96}  &   18.04 \scriptsize{$\pm$  1.28}  & \textbf{92.74} \scriptsize{$\pm$ 0.77} & \underline{71.64} \scriptsize{$\pm$ 0.55} \\
    
  & +MI  &  70.95 \scriptsize{$\pm$ 1.16}  &  29.21 \scriptsize{$\pm$ 1.33} & 	   10.54 \scriptsize{$\pm$ 0.33} & 88.75 \scriptsize{$\pm$ 1.36} & 66.88 \scriptsize{$\pm$ 0.69} \\

  & +SimCLR  &  74.26 \scriptsize{$\pm$ 0.43}  &  31.26 \scriptsize{$\pm$ 0.15}  &   13.93 \scriptsize{$\pm$ 0.48} & 89.12 \scriptsize{$\pm$ 1.43}  & 66.90 \scriptsize{$\pm$ 1.56} \\

 & +BYOL  & 73.56 \scriptsize{$\pm$ 0.21}  &  38.74 \scriptsize{$\pm$ 2.43} & 	   \underline{18.43} \scriptsize{$\pm$ 0.77} & 91.17 \scriptsize{$\pm$ 0.42} & 66.87 \scriptsize{$\pm$ 0.55} \\

 & +DINO  &  72.14 \scriptsize{$\pm$ 0.70}  &  35.33 \scriptsize{$\pm$ 0.78} & 	   14.19 \scriptsize{$\pm$  0.69}  & 88.08 \scriptsize{$\pm$ 0.41} & 62.41 \scriptsize{$\pm$ 3.23} \\

 & +Barlow Twins  &  73.23 \scriptsize{$\pm$ 1.28}  &  35.23 \scriptsize{$\pm$ 1.73}   &   18.05 \scriptsize{$\pm$  0.26} & 91.08 \scriptsize{$\pm$ 0.67} & 67.03 \scriptsize{$\pm$ 1.00}  \\
\bottomrule
\end{tabular}
\end{table}

\section{Results}
Table \ref{tab:main} provides the effect of various consistency regularizers added to ER in different CL scenarios. Our results for the Class-IL scenario show that:
\begin{enumerate}[\hspace{24pt}(i)]
  \item Across all datasets, regularizing consistency in predictions is almost always better than vanilla ER. Consistency helps in preserving the rich information about the previous tasks better thereby further mitigating forgetting.
  \item The improvement in performance is more significant in more complex datasets. For example in S-TinyImageNet, the final Top-1 Accuracy with +$L_{inf}$ is almost double that of vanilla ER for the buffer size of 500.
  \item Reinforcing our earlier hypothesis, consistency regularization in ER can indeed be cast as a self-supervised pretext task enabling the use of state-of-the-art SSL methods in CL. SSL methods assist ER in mitigating forgetting in the majority of the experiments.
  \item Among SSL regularizers, cross-correlation-based and MSE-based methods perform comparably well. In Barlow Twins, on-diagonal elements in the cross-correlation matrix carry a higher weight resulting in a stronger form of consistency among predictions. Maximizing cosine similarity between predictions using BYOL in our framework translates to minimizing squared error when the predictions are $L_2$ normalized (Appendix \ref{byol}). Therefore, +BYOL's performance closely follows that of +MSE's.
  \item All regularizers perform equally well on S-CIFAR-10. However, approximate methods (e.g., +MI, +SimCLR, +DINO) struggle to keep up with the rest of regularizers as number of classes increases. These methods do not fully mimic the past behaviour but approximate it by bringing the corresponding predictions closer in the latent space creating a knowledge gap. This knowledge gap is especially more in high-dimensional spaces leading to poor performance. We argue that stricter consistency constraints such as +$L_p$ and +MSE are more successful to close this knowledge gap.
  \item Similarly, +KL-Div fails miserably in higher dimensional spaces. We think this is due to the loss of information occurring in probability space due to the squashing function (e.g., softmax) \citep{buzzega2020dark}.
    \item The Chebyshev distance +$L_{inf}$, a simple form of regularization, is surprisingly one of the top performers in Class-IL scenario across all datasets. As ER has a high task recency bias (see Section \ref{task_bias}), the current predictions are skewed towards current task logits. However, the past predictions are skewed towards corresponding past task logits. As +$L_{inf}$ considers only the greatest difference along any coordinate dimension, we suspect that maximum difference lies either along the current task logits or corresponding past task logits. Therefore, +$L_{inf}$ penalizes the task recency bias and aligns past and current predictions along past task logits' dimensions in subsequent training iterations. Due to this, +$L_{inf}$ is able to retain its top-notch performance even in high-dimensional spaces. 
\end{enumerate}

The benefits of maintaining consistency in predictions is not only limited to the Class-IL scenario. 
% Domain-IL is more challenging than Class-IL. 
R-MNIST under Domain-IL scenario requires the CL model to classify all MNIST digits for 20 subsequent tasks, for rotated images by a random angle in the interval $[0, \pi)$. In order to perform well, models need to enhance their positive forward transfer all the while lessening interference. 
Our earlier inferences can well be extrapolated here: using consistency regularization is almost always beneficial to preserve rich information about the previous tasks thereby enhancing positive forward transfer. +$L_p$ family of regularizers are top performers while approximate methods doing slightly worse. Among SSL algorithms, +BYOL and +Barlow Twins perform equally well compared to top performers. 

MNIST-360 is a new protocol developed to address the general continual learning  desiderata \citep{buzzega2020dark}. It models a stream of two consecutive MNIST digits (\{0,1\}, \{1,2\}, ..., \{7,8\}) with increasing rotation applied in subsequent batches. It is worth noting that such a setting offers both a sharp shift in terms of classes and a smooth shift in terms of rotation. MNIST-360 involves recurring rotated digits in subsequent sequences which makes the transfer of knowledge from previous occurrences important. As is the case in earlier scenarios, +$L_p$ consistency regularization significantly augments the ER by effectively preserving the previous knowledge. The difference in performance between stricter and approximate consistency regularizers in this scenario is due to the additional complexity added by the rotation. +$L_p$ regularizers leverage rich information about the past better when revisiting the previous tasks and outperform the ER across different buffer sizes. 

It is indeed clear from the above analyses under different CL scenarios and datasets that consistency regularization plays a prominent role in alleviating the catastrophic forgetting in ER. Enforcing consistency in predictions separated through time helps in preserving rich information about the previous tasks better thereby helping to further mitigate the catastrophic forgetting in CL.

\begin{table}[t]
% \centering
    \begin{minipage}{.49\linewidth}
        \centering
        \caption{Comparison of Top-1 Accuracy ($\%$) on S-CIFAR-100 under longer task sequences. Best result is in bold and the second best is underlined.}
        \label{tab:cifar100}
            \begin{tabular}{l|ccc}
            \toprule
            Method & 5 tasks & 10 tasks & 20 tasks  \\ 
            \midrule
            ER	& 28.02\scriptsize{$\pm$0.31} & 21.49\scriptsize{$\pm$0.47} & 16.52\scriptsize{$\pm$0.86}\\
            +$L_{1}$ & 	 40.40\scriptsize{$\pm$1.31} &   35.75\scriptsize{$\pm$0.25}  &  	  \textbf{25.30}\scriptsize{$\pm$6.40 }\\
             +$L_{2}$ & 	 40.92\scriptsize{$\pm$0.95} &   34.32\scriptsize{$\pm$0.49} &	 23.51\scriptsize{$\pm$4.71 }\\
             +$L_{inf}$ & 		\textbf{42.15}\scriptsize{$\pm$1.88} &   \textbf{38.03}\scriptsize{$\pm$0.81} &	  \underline{25.05}\scriptsize{$\pm$7.00}\\
             +KL-Div & 	27.91\scriptsize{$\pm$0.98} &   	22.38\scriptsize{$\pm$1.13} &	  15.32\scriptsize{$\pm$5.40 }\\
             +MSE & \underline{41.40}\scriptsize{$\pm$0.96} &    	\underline{36.20}\scriptsize{$\pm$0.52} &	  22.25\scriptsize{$\pm$5.87 }\\
            +MI	& 29.21\scriptsize{$\pm$1.33} & 22.73\scriptsize{$\pm$1.39}& 17.40\scriptsize{$\pm$2.93}	\\
            +SimCLR	& 31.26\scriptsize{$\pm$0.15} & 25.89\scriptsize{$\pm$0.99}& 19.48\scriptsize{$\pm$4.55}	\\
            +BYOL	& 38.74\scriptsize{$\pm$2.43} & 35.84\scriptsize{$\pm$0.96} & 22.99\scriptsize{$\pm$4.42}\\
            +DINO	& 35.33\scriptsize{$\pm$0.78} & 30.24\scriptsize{$\pm$1.43} & 20.49\scriptsize{$\pm$4.37}	\\
            +Barlow Twins	& 35.23\scriptsize{$\pm$1.73} & 31.07\scriptsize{$\pm$1.50} & 18.79\scriptsize{$\pm$5.94} \\
            \bottomrule
            \end{tabular}
    \end{minipage}%
    \hspace{0.05\textwidth}%
    % \vline
    % \hspace{0.001\textwidth}
    \begin{minipage}{.45\linewidth}
        \caption{Comparison of Top-1 Accuracy ($\%$) on S-CIFAR-10 under extremely low-buffer sizes. Best result is in bold and the second best is underlined.}
        \label{tab:sample_complexity}
            \begin{tabular}{cccc}
                \toprule
                 10 & 20 & 50 & 100 \\ 
                \midrule
            22.20\scriptsize{$\pm$1.36} & 26.69\scriptsize{$\pm$1.21}& 32.51\scriptsize{$\pm$1.77} & 41.10\scriptsize{$\pm$1.10}\\
            28.55 \scriptsize{$\pm$2.58} &   \underline{34.77}\scriptsize{$\pm$3.59}&  \underline{47.42}\scriptsize{$\pm$2.20} &  \textbf{57.92}\scriptsize{$\pm$1.47}\\ 
             26.00\scriptsize{$\pm$6.01} &  33.97\scriptsize{$\pm$5.30}&  45.86\scriptsize{$\pm$3.14} &  55.69\scriptsize{$\pm$1.00}\\ 
             27.91\scriptsize{$\pm$3.19} &  33.32\scriptsize{$\pm$5.19}& \textbf{47.84}\scriptsize{$\pm$2.39} & \underline{57.67}\scriptsize{$\pm$0.60}\\ 
             21.93\scriptsize{$\pm$0.95} &  25.06\scriptsize{$\pm$1.39}&  35.74\scriptsize{$\pm$2.50} &  47.93\scriptsize{$\pm$2.04}\\
              \textbf{30.21}\scriptsize{$\pm$1.78} &   32.55\scriptsize{$\pm$1.90}&  45.27\scriptsize{$\pm$2.34} & 56.09\scriptsize{$\pm$0.87}\\
             23.51\scriptsize{$\pm$0.34} & 29.10\scriptsize{$\pm$1.48}& 36.13\scriptsize{$\pm$3.48}& 50.19\scriptsize{$\pm$0.74} \\ 
             27.72\scriptsize{$\pm$2.63} & 28.54\scriptsize{$\pm$3.02}& 42.61\scriptsize{$\pm$2.91}& 51.91\scriptsize{$\pm$0.19}\\
            \underline{29.04}\scriptsize{$\pm$1.57}& \textbf{35.40}\scriptsize{$\pm$2.59}& 43.62\scriptsize{$\pm$3.40}& 55.73\scriptsize{$\pm$2.52}\\
            24.96\scriptsize{$\pm$0.98}& 34.36\scriptsize{$\pm$1.14}& 39.65\scriptsize{$\pm$1.48}& 52.60\scriptsize{$\pm$1.33}	\\
            25.24\scriptsize{$\pm$4.40}& 33.54\scriptsize{$\pm$3.43}& 42.92\scriptsize{$\pm$3.56}& 54.48\scriptsize{$\pm$1.55} \\
        \bottomrule
        \end{tabular}
    \end{minipage} 
\end{table}

\section{Model Analysis}

\subsection{Effect of longer task sequences and low-buffer regimes}
Catastrophic forgetting worsens as the number of tasks in a sequence increases, referred to as long-term catastrophic forgetting \citep{9354016}. The number of samples in the buffer representing each previous task drastically reduces in longer task sequences resulting in poor performance for ER. As the number of classes are low in S-CIFAR-10, we consider S-CIFAR-100 for this analysis. Table \ref{tab:cifar100} shows the performance of models with 5, 10, and 20 task sequences on S-CIFAR-100 with a fixed buffer size of 500. Our analysis of Class-IL can straightforwardly be extrapolated here: Except +KL-Div, all regularizers augment ER in preserving rich previous task information in longer task sequences. As was the case earlier, stricter consistency constraints (e.g., +$L_p$, +MSE) enable better performance than approximate methods. +$L_{inf}$ brings a relative improvement of at least $50\%$ across all task sequences. Although SSL methods lag behind +$L_{inf}$, they still outperform ER with a good margin. 

% The performance of ER directly correlates with the buffer size, as evidenced in our empirical results. Therefore, it is important to see whether consistency regularization can benefit ER in low-buffer regimes. 
Table \ref{tab:sample_complexity} presents the performance under extremely low-buffer sizes. As S-CIFAR-10 consists of 10 classes, extreme low-buffer regimes (e.g. 10, 20) may not even have representative samples from all classes in the buffer due to Reservoir Sampling. Under low buffer regimes, ER learns discriminative features specific to buffered samples thereby inaccurately approximating the previous task data distributions. These learned features are not representative of the past tasks and do not generalize well. Since soft-targets capture additional information with regard to the similarity structure of the data, 
regularizing consistency in predictions of soft-targets preserves the previous task information better compared to vanilla ER.
% Even under extreme low buffer regimes, regularizing consistency in predictions throughout the CL training helps mitigate catastrophic forgetting further.  

Consistency regularization brings a significant performance boost to ER even under extremely low-buffer sizes and longer task sequences by mitigating catastrophic forgetting all the while reducing the reliance on buffer size. Therefore, consistency regularization is the need of the hour for rehearsal-based methods to preserve rich information about the previous tasks better.

% \subsection{Sample complexity}
% The performance of ER-based methods directly correlates with the buffer size, as evidenced in our empirical results. 
% % However, the performance takes a hit in longer task sequences. For a fixed memory budget, there will be less number of samples representing previous tasks in the memory buffer in longer task sequences. Also, a
% As the number of tasks grow in CL, scalability will be an issue if buffer grows in size commensurate with the number of tasks.  Therefore, it is quintessential to develop ER-based methods that are sample efficient and learn generalizable features across tasks. Table \ref{tab:sample_complexity} presents the buffered sample complexity of different CL models. Models are trained with constrained buffer sizes to evaluate their ability to mitigate catastrophic forgetting. Reinforcing our earlier hypothesis, using any kind of regularization helps in further mitigating catastrophic forgetting in CL. For a given buffer size, consistency regularization enables CL model to preserve more information about past tasks, thereby improving stability and consequently reducing forgetting. Akin to ER, performance of models within our framework improve as the buffer grows in size indicating the suitability of our framework across all low-buffer regimes. 

\begin{figure}[t]
\centering
  \includegraphics[width=1\linewidth]{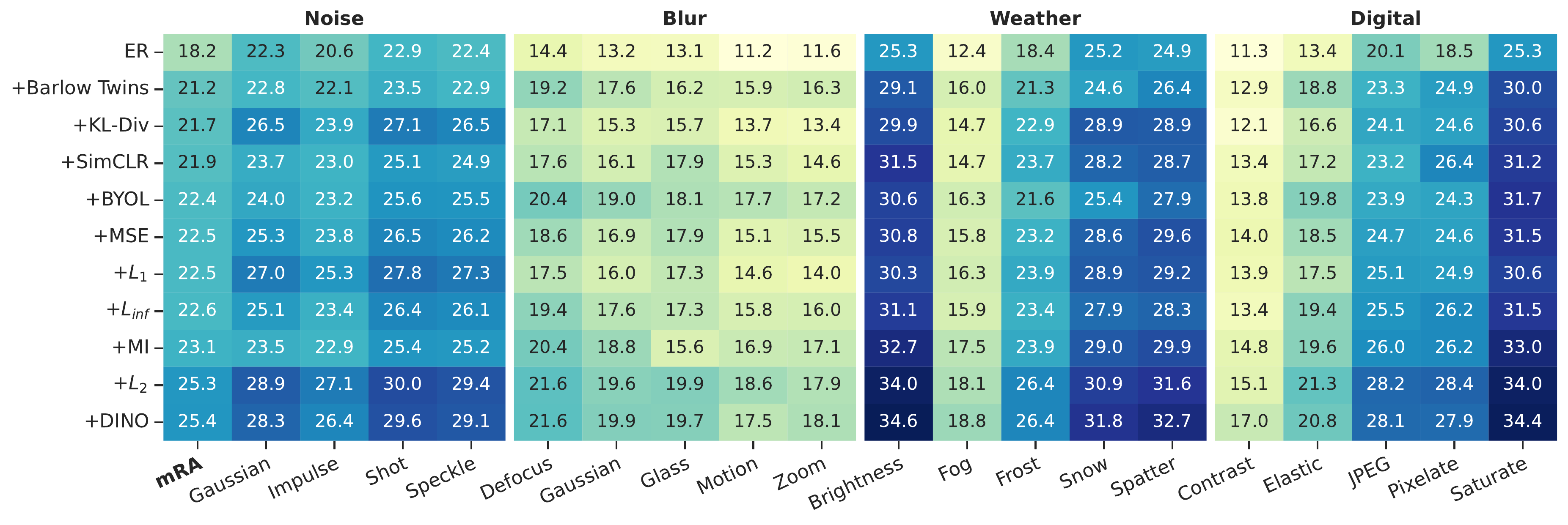}
  \caption{Robustness to common natural input corruptions on noise, blur, weather, and digital categories as Top-1 Accuracy (\%). Methods are sorted in the ascending manner based on their mean robust accuracy (mRA).}
  \label{fig:robustness}
\end{figure}

\subsection{Robustness to image corruptions}\label{robust}
% Distributional shift, where the training distribution differs from the test distribution
Distributional shift often influenced by weather and illumination changes is widely prevalent in the real world. DNNs have been shown to be sensitive to distributional shifts such as common input corruptions \citep{hendrycks2018benchmarking}. Consequently, CL agents operating in high-stakes applications such as autonomous driving need to be robust to a variety of such distributional shifts. In this section, we evaluate the effect of added consistency regularizers to ER against common synthetic image corruptions using CIFAR-10-C \citep{hendrycks2018benchmarking}. CL models are trained on S-CIFAR-10 (buffer size 500) and evaluated on the CIFAR-10-C validation set. Figure \ref{fig:robustness} presents the robust accuracy of CL models on 15 different out-of-distribution datasets under noise, blur, weather, and digital categories.
ER is least robust while all regularizers on top of ER are more robust to induced natural corruptions. As ER relies solely on hard targets to approximate the past behaviour, it learns features specific to buffered samples. Therefore, learned representations do not generalize well to out-of-distribution datasets. As past predictions hold rich information about previous tasks, the regularization scheme enforces ER to learn generalizable features making it more robust to distributional shifts. 
% All regularizers improve ER's robustness to input corruptions. 

\begin{figure}[tb]
\centering
  \includegraphics[width=1\linewidth]{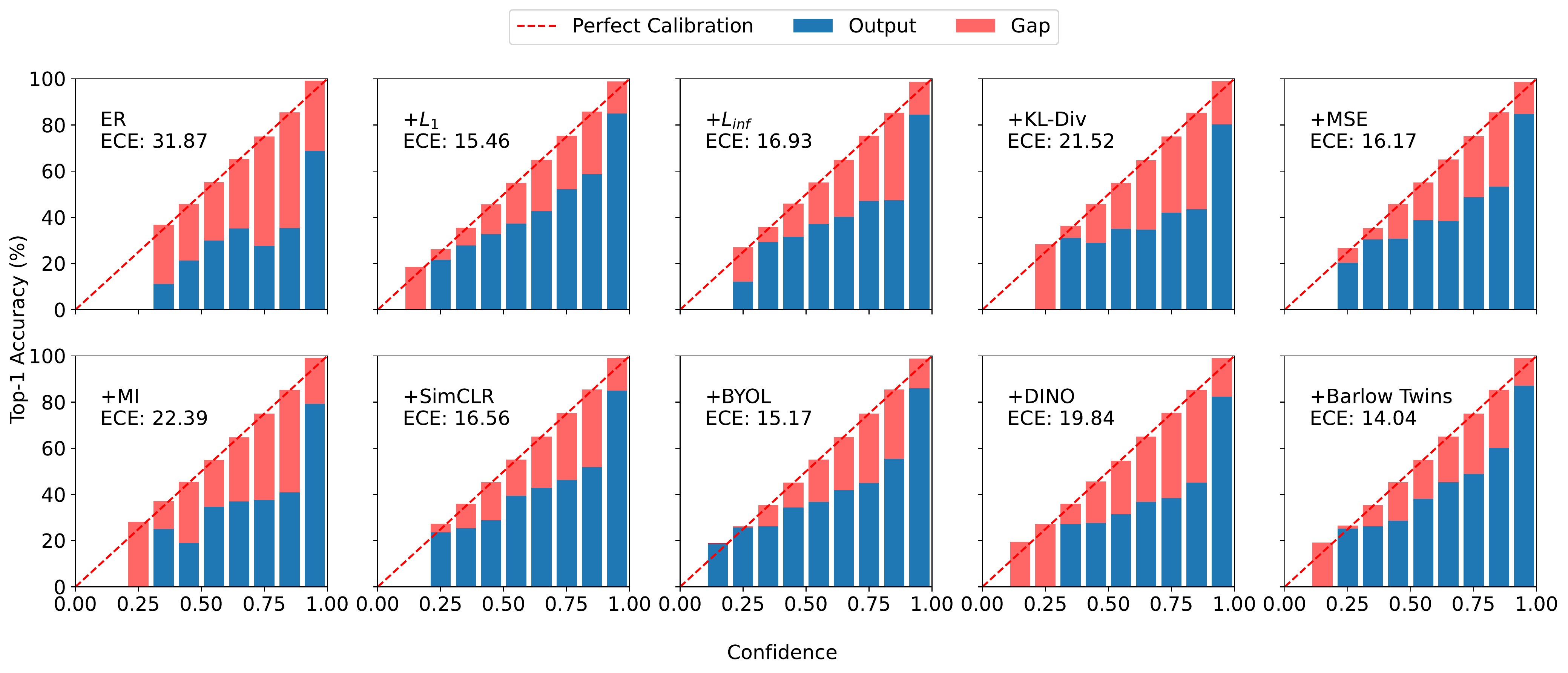}
  \caption{Confidence estimates and corresponding Expected Calibration Error (ECE) of CL models trained on S-CIFAR-10 with buffer size 500. Lower ECE indicates more reliability.}
  \label{fig:calib}
\end{figure}

\subsection{Confidence calibration}\label{sec:calib}
A well-calibrated model improving reliability is quintessential for safety-critical CL systems. Confidence calibration entails the problem of predicting probability estimates representative of the true correctness likelihood. Miscalibration can be defined as the difference in expectation between confidence and accuracy \citep{guo2017calibration}. Expected Calibration Error (ECE) (Appendix  \ref{ece_computation}) approximates the miscalibration in classification by partitioning the predictions into equal-sized bins and computes the difference between the weighted average of the bins’ accuracy and confidence \citep{naeini2015obtaining}. A lower ECE value represents better calibration in underlying models. Figure \ref{fig:calib} shows ECE along with a reliability diagram on S-CIFAR-10 using a calibration framework \citep{kuppers2020multivariate}. ER is highly miscalibrated and far more overconfident than models trained using a regularizer. On the other hand, all added regularizers to ER improve the model calibration. Stricter consistency constraints have better calibration, even reducing the ECE score by half in most cases. Consistency regularization ensures that predicted softmax scores are better indicators of the actual likelihood of a correct prediction thereby improving the reliability of ER in real-world applications.

% Regularizers which impose stronger penalty (e.g. $L_p$, barlow twins) have a better calibration while others (e.g. MI, KL-Div, DINO) are relatively miscalibrated due to loose form consistency.

% We argue that consistency regularization enables learning rich discriminative features quintessential for well-calibrated models. 
% In addition to improvement in natural accuracy, our framework provides an intrinsic way of improving calibration and thus reliability in safety-critical CL systems. 

\subsection{Task recency bias}\label{task_bias}
\begin{figure}[tb]
\centering
  \includegraphics[width=1\linewidth]{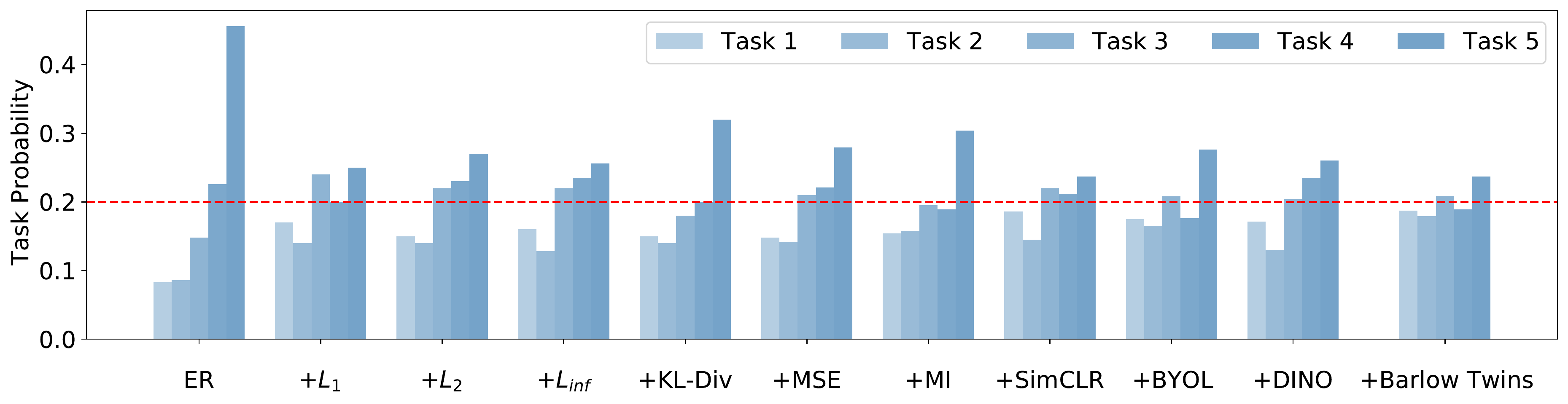}
  \caption{Average task probabilities of CL models trained on
CIFAR-10 with 500 buffer size. Within each bar group, right most
bar indicates the most recent task. }
  \label{fig:task_recency_bias}
\end{figure}

Replay-based methods are prone to task recency bias - the tendency of a CL model to be biased towards classes from the most recent tasks \citep{masana2020class, hou2019learning}. Specifically, the model sees only a few samples from the old tasks while aplenty from the most recent task, leading to decisions biased towards new classes and the confusion among old classes. 
% Explicit techniques such as data resampling \citep{he2008adasyn} and cost-sensitive learning \citep{huang2016learning} have been used in the literature to address learning under class-imbalance setting. 
Figure \ref{fig:task_recency_bias} presents the the normalized probabilities of each task of a S-CIFAR-10 trained model, computed by averaging probabilities of all samples belonging to the associated classes in a Class-IL setting. The predictions in vanilla  ER are biased mostly towards recent tasks, with most recent task being almost $4X$ as much as the first task. Therefore, the predictions stored in the buffer are completely biased towards their corresponding task logits. Since consistency regularization penalizes any deviation in predictions, task recency bias is mitigated as a by-product.

% Barlow Twins has one of most evenly distributed task probabilities while KL-Div and MI have relatively high task recency bias. 
% Once stored in the buffer, these predictions are aligned towards their corresponding task logits. Consistency regularization penalizes any deviation in predictions thereby evenly distributing task probabilities. 

% However, task probabilities are more evenly distributed for methods with a regularizer reducing the recency bias. Essentially, consistency regularization tries to match current and past predictions,

% Consistency regularizers penalize learning superficial features thereby incentivizing more evenly distributed task predictions all the while retaining accuracy on recent tasks.  
% in our framework work to improve affinity towards positive pairs separated through time, thereby reducing task recency bias all the while retaining accuracy on recent tasks.  % Mutual  information in Equation \ref{eqn:mi} expands to  $I(z_\theta, z_r)  = H(z_\theta) - H(z_\theta|z_r)$. For buffered samples, $H(z_\theta)$ is fixed. Hence, maximizing MI trades-off minimizing the conditional entropy $H(z_\theta|z_r)$. The smallest value of conditional entropy is obtained when class assignments are exactly predictable from each other. This happens when task probabilities are evenly distributed across tasks. 

% \textcolor{blue}{In addition to alleviating catastrophic forgetting, consistency regularization enhances robustness to natural corruptions and model calibration while simultaneously reducing task recency bias. Therefore, regularizing consistency in predictions is crucial for wholesome improvement of ER.}

\section{Conclusion}
We provided a comprehensive study on the role of consistency regularization under a common ER framework across various CL scenarios and datasets. We also studied the state-of-the-art SSL algorithms as regularizers by interpreting the regularization of consistency in predictions as a self-supervised pretext task. Consistency regularization augments ER in preserving rich information about the previous tasks better thereby mitigating catastrophic forgetting further. Even under extremely low-buffer sizes and longer task sequences, it yields discernible performance improvement. Stricter consistency constraints are better in mitigating forgetting compared to other approximate methods considered in this study. Minkowski distance functions, extremely simple and efficient, are surprisingly effective in alleviating catastrophic forgetting. Additionally, regularizing consistency in predictions leads to more robust and well-calibrated model. It also mitigates the task recency bias thereby leading to more evenly distributed task prediction probabilities. Therefore, regularizing consistency in predictions is crucial for wholesome improvement of ER. We believe that the representation-replay and generative-replay based methods can also benefit from the analysis conducted in this paper. 
% We conclude that consistency regularization is a must-have for rehearsal-based methods. 
We hope that this study further spurs research in advancing CL towards sample efficient experience replay. 
\bibliography{egbib}

\begin{thebibliography}{60}
\providecommand{\natexlab}[1]{#1}
\providecommand{\url}[1]{\texttt{#1}}
\expandafter\ifx\csname urlstyle\endcsname\relax
  \providecommand{\doi}[1]{doi: #1}\else
  \providecommand{\doi}{doi: \begingroup \urlstyle{rm}\Url}\fi

\bibitem[Arani et~al.(2022)Arani, Sarfraz, and Zonooz]{arani2022learning}
Elahe Arani, Fahad Sarfraz, and Bahram Zonooz.
\newblock Learning fast, learning slow: A general continual learning method
  based on complementary learning system.
\newblock In \emph{International Conference on Learning Representations}, 2022.

\bibitem[Bachman et~al.(2014)Bachman, Alsharif, and
  Precup]{bachman2014learning}
Philip Bachman, Ouais Alsharif, and Doina Precup.
\newblock Learning with pseudo-ensembles.
\newblock \emph{Advances in neural information processing systems}, 27, 2014.

\bibitem[Benjamin et~al.(2018)Benjamin, Rolnick, and Kording]{fdr}
Ari Benjamin, David Rolnick, and Konrad Kording.
\newblock Measuring and regularizing networks in function space.
\newblock In \emph{International Conference on Learning Representations}, 2018.

\bibitem[Buzzega et~al.(2020)Buzzega, Boschini, Porrello, Abati, and
  Calderara]{buzzega2020dark}
Pietro Buzzega, Matteo Boschini, Angelo Porrello, Davide Abati, and Simone
  Calderara.
\newblock Dark experience for general continual learning: a strong, simple
  baseline.
\newblock In \emph{34th Conference on Neural Information Processing Systems
  (NeurIPS 2020)}, 2020.

\bibitem[Caccia et~al.(2020)Caccia, Belilovsky, Caccia, and
  Pineau]{caccia2020online}
Lucas Caccia, Eugene Belilovsky, Massimo Caccia, and Joelle Pineau.
\newblock Online learned continual compression with adaptive quantization
  modules.
\newblock In \emph{International Conference on Machine Learning}, pp.\
  1240--1250. PMLR, 2020.

\bibitem[Caron et~al.(2021)Caron, Touvron, Misra, J\'egou, Mairal, Bojanowski,
  and Joulin]{Caron_2021_ICCV}
Mathilde Caron, Hugo Touvron, Ishan Misra, Herv\'e J\'egou, Julien Mairal,
  Piotr Bojanowski, and Armand Joulin.
\newblock Emerging properties in self-supervised vision transformers.
\newblock In \emph{Proceedings of the IEEE/CVF International Conference on
  Computer Vision (ICCV)}, pp.\  9650--9660, October 2021.

\bibitem[Chaudhry et~al.(2019)Chaudhry, Ranzato, Rohrbach, and Elhoseiny]{agem}
Arslan Chaudhry, Marc’Aurelio Ranzato, Marcus Rohrbach, and Mohamed
  Elhoseiny.
\newblock Efficient lifelong learning with a-gem.
\newblock In \emph{ICLR}, 2019.

\bibitem[Chen et~al.(2020)Chen, Kornblith, Norouzi, and Hinton]{chen2020simple}
Ting Chen, Simon Kornblith, Mohammad Norouzi, and Geoffrey Hinton.
\newblock A simple framework for contrastive learning of visual
  representations.
\newblock In \emph{International conference on machine learning}, pp.\
  1597--1607. PMLR, 2020.

\bibitem[Chen \& He(2021)Chen and He]{chen2021exploring}
Xinlei Chen and Kaiming He.
\newblock Exploring simple siamese representation learning.
\newblock In \emph{Proceedings of the IEEE/CVF Conference on Computer Vision
  and Pattern Recognition}, pp.\  15750--15758, 2021.

\bibitem[Davari et~al.(2022)Davari, Asadi, Mudur, Aljundi, and
  Belilovsky]{davari2022probing}
MohammadReza Davari, Nader Asadi, Sudhir Mudur, Rahaf Aljundi, and Eugene
  Belilovsky.
\newblock Probing representation forgetting in supervised and unsupervised
  continual learning.
\newblock \emph{arXiv preprint arXiv:2203.13381}, 2022.

\bibitem[Farquhar \& Gal(2018)Farquhar and Gal]{farquhar2018towards}
Sebastian Farquhar and Yarin Gal.
\newblock Towards robust evaluations of continual learning.
\newblock \emph{arXiv preprint arXiv:1805.09733}, 2018.

\bibitem[Fini et~al.(2021)Fini, da~Costa, Alameda-Pineda, Ricci, Alahari, and
  Mairal]{fini2021self}
Enrico Fini, Victor G~Turrisi da~Costa, Xavier Alameda-Pineda, Elisa Ricci,
  Karteek Alahari, and Julien Mairal.
\newblock Self-supervised models are continual learners.
\newblock \emph{arXiv preprint arXiv:2112.04215}, 2021.

\bibitem[Goodfellow et~al.(2013)Goodfellow, Mirza, Xiao, Courville, and
  Bengio]{goodfellow2013empirical}
Ian~J Goodfellow, Mehdi Mirza, Da~Xiao, Aaron Courville, and Yoshua Bengio.
\newblock An empirical investigation of catastrophic forgetting in
  gradient-based neural networks.
\newblock \emph{arXiv preprint arXiv:1312.6211}, 2013.

\bibitem[Grill et~al.(2020)Grill, Strub, Altch{\'e}, Tallec, Richemond,
  Buchatskaya, Doersch, Pires, Guo, Azar, et~al.]{grill2020bootstrap}
Jean-Bastien Grill, Florian Strub, Florent Altch{\'e}, Corentin Tallec, Pierre
  Richemond, Elena Buchatskaya, Carl Doersch, Bernardo Pires, Zhaohan Guo,
  Mohammad Azar, et~al.
\newblock Bootstrap your own latent: A new approach to self-supervised
  learning.
\newblock In \emph{Neural Information Processing Systems}, 2020.

\bibitem[Grossberg(1982)]{Grossberg1982}
Stephen Grossberg.
\newblock How does a brain build a cognitive code?
\newblock \emph{Studies of mind and brain}, pp.\  1--52, 1982.

\bibitem[Guo et~al.(2017)Guo, Pleiss, Sun, and Weinberger]{guo2017calibration}
Chuan Guo, Geoff Pleiss, Yu~Sun, and Kilian~Q Weinberger.
\newblock On calibration of modern neural networks.
\newblock In \emph{International Conference on Machine Learning}, pp.\
  1321--1330. PMLR, 2017.

\bibitem[Gutmann \& Hyv{\"a}rinen(2012)Gutmann and
  Hyv{\"a}rinen]{gutmann2012noise}
Michael~U Gutmann and Aapo Hyv{\"a}rinen.
\newblock Noise-contrastive estimation of unnormalized statistical models, with
  applications to natural image statistics.
\newblock \emph{Journal of Machine Learning Research}, 13\penalty0 (2), 2012.

\bibitem[Hassabis et~al.(2017)Hassabis, Kumaran, Summerfield, and
  Botvinick]{HASSABIS2017245}
Demis Hassabis, Dharshan Kumaran, Christopher Summerfield, and Matthew
  Botvinick.
\newblock Neuroscience-inspired artificial intelligence.
\newblock \emph{Neuron}, 95\penalty0 (2):\penalty0 245--258, 2017.
\newblock ISSN 0896-6273.
\newblock \doi{https://doi.org/10.1016/j.neuron.2017.06.011}.

\bibitem[He et~al.(2016)He, Zhang, Ren, and Sun]{he2016deep}
Kaiming He, Xiangyu Zhang, Shaoqing Ren, and Jian Sun.
\newblock Deep residual learning for image recognition.
\newblock In \emph{Proceedings of the IEEE conference on computer vision and
  pattern recognition}, pp.\  770--778, 2016.

\bibitem[He et~al.(2020)He, Fan, Wu, Xie, and Girshick]{he2020momentum}
Kaiming He, Haoqi Fan, Yuxin Wu, Saining Xie, and Ross Girshick.
\newblock Momentum contrast for unsupervised visual representation learning.
\newblock In \emph{Proceedings of the IEEE/CVF Conference on Computer Vision
  and Pattern Recognition}, pp.\  9729--9738, 2020.

\bibitem[Hendrycks \& Dietterich(2018)Hendrycks and
  Dietterich]{hendrycks2018benchmarking}
Dan Hendrycks and Thomas Dietterich.
\newblock Benchmarking neural network robustness to common corruptions and
  perturbations.
\newblock In \emph{International Conference on Learning Representations}, 2018.

\bibitem[Hinton et~al.(2015)Hinton, Vinyals, and Dean]{hinton2015}
Geoffrey Hinton, Oriol Vinyals, and Jeffrey Dean.
\newblock Distilling the knowledge in a neural network.
\newblock In \emph{NIPS Deep Learning and Representation Learning Workshop},
  2015.

\bibitem[Hou et~al.(2019)Hou, Pan, Loy, Wang, and Lin]{hou2019learning}
Saihui Hou, Xinyu Pan, Chen~Change Loy, Zilei Wang, and Dahua Lin.
\newblock Learning a unified classifier incrementally via rebalancing.
\newblock In \emph{Proceedings of the IEEE/CVF Conference on Computer Vision
  and Pattern Recognition}, pp.\  831--839, 2019.

\bibitem[Hua et~al.(2021)Hua, Wang, Xue, Ren, Wang, and Zhao]{hua2021feature}
Tianyu Hua, Wenxiao Wang, Zihui Xue, Sucheng Ren, Yue Wang, and Hang Zhao.
\newblock On feature decorrelation in self-supervised learning.
\newblock In \emph{Proceedings of the IEEE/CVF International Conference on
  Computer Vision}, pp.\  9598--9608, 2021.

\bibitem[Kirkpatrick et~al.(2017)Kirkpatrick, Pascanu, Rabinowitz, Veness,
  Desjardins, Rusu, Milan, Quan, Ramalho, Grabska-Barwinska,
  et~al.]{kirkpatrick2017overcoming}
James Kirkpatrick, Razvan Pascanu, Neil Rabinowitz, Joel Veness, Guillaume
  Desjardins, Andrei~A Rusu, Kieran Milan, John Quan, Tiago Ramalho, Agnieszka
  Grabska-Barwinska, et~al.
\newblock Overcoming catastrophic forgetting in neural networks.
\newblock \emph{Proceedings of the national academy of sciences}, 114\penalty0
  (13):\penalty0 3521--3526, 2017.

\bibitem[Krizhevsky(2009)]{cifar10}
A.~Krizhevsky.
\newblock Learning multiple layers of features from tiny images.
\newblock 2009.

\bibitem[Kudugunta et~al.(2019)Kudugunta, Bapna, Caswell, Arivazhagan, and
  Firat]{kudugunta2019investigating}
Sneha~Reddy Kudugunta, Ankur Bapna, Isaac Caswell, Naveen Arivazhagan, and
  Orhan Firat.
\newblock Investigating multilingual nmt representations at scale.
\newblock \emph{arXiv preprint arXiv:1909.02197}, 2019.

\bibitem[Kumaran et~al.(2016)Kumaran, Hassabis, and
  McClelland]{kumaran2016learning}
Dharshan Kumaran, Demis Hassabis, and James~L McClelland.
\newblock What learning systems do intelligent agents need? complementary
  learning systems theory updated.
\newblock \emph{Trends in cognitive sciences}, 20\penalty0 (7):\penalty0
  512--534, 2016.

\bibitem[Kuppers et~al.(2020)Kuppers, Kronenberger, Shantia, and
  Haselhoff]{kuppers2020multivariate}
Fabian Kuppers, Jan Kronenberger, Amirhossein Shantia, and Anselm Haselhoff.
\newblock Multivariate confidence calibration for object detection.
\newblock In \emph{Proceedings of the IEEE/CVF Conference on Computer Vision
  and Pattern Recognition Workshops}, pp.\  326--327, 2020.

\bibitem[Le \& Yang(2015)Le and Yang]{tinyimagenet}
Ya~Le and X.~Yang.
\newblock Tiny imagenet visual recognition challenge.
\newblock 2015.

\bibitem[Lecun et~al.(1998)Lecun, Bottou, Bengio, and Haffner]{mnist}
Y.~Lecun, L.~Bottou, Y.~Bengio, and P.~Haffner.
\newblock Gradient-based learning applied to document recognition.
\newblock \emph{Proceedings of the IEEE}, 86\penalty0 (11):\penalty0
  2278--2324, 1998.
\newblock \doi{10.1109/5.726791}.

\bibitem[Li \& Hoiem(2017)Li and Hoiem]{lwf}
Zhizhong Li and Derek Hoiem.
\newblock Learning without forgetting.
\newblock \emph{IEEE transactions on pattern analysis and machine
  intelligence}, 40\penalty0 (12):\penalty0 2935--2947, 2017.

\bibitem[Lopez-Paz \& Ranzato(2017)Lopez-Paz and Ranzato]{gem}
David Lopez-Paz and Marc'Aurelio Ranzato.
\newblock Gradient episodic memory for continual learning.
\newblock \emph{Advances in neural information processing systems},
  30:\penalty0 6467--6476, 2017.

\bibitem[Madaan et~al.(2021)Madaan, Yoon, Li, Liu, and
  Hwang]{madaan2021representational}
Divyam Madaan, Jaehong Yoon, Yuanchun Li, Yunxin Liu, and Sung~Ju Hwang.
\newblock Representational continuity for unsupervised continual learning.
\newblock In \emph{International Conference on Learning Representations}, 2021.

\bibitem[Masana et~al.(2020)Masana, Liu, Twardowski, Menta, Bagdanov, and
  van~de Weijer]{masana2020class}
Marc Masana, Xialei Liu, Bartlomiej Twardowski, Mikel Menta, Andrew~D Bagdanov,
  and Joost van~de Weijer.
\newblock Class-incremental learning: survey and performance evaluation on
  image classification.
\newblock \emph{arXiv preprint arXiv:2010.15277}, 2020.

\bibitem[McCloskey \& Cohen(1989)McCloskey and
  Cohen]{mccloskey1989catastrophic}
Michael McCloskey and Neal~J Cohen.
\newblock Catastrophic interference in connectionist networks: The sequential
  learning problem.
\newblock In \emph{Psychology of learning and motivation}, volume~24, pp.\
  109--165. Elsevier, 1989.

\bibitem[Mermillod et~al.(2013)Mermillod, Bugaiska, and Bonin]{Bugaiska}
Martial Mermillod, Aur{\'e}lia Bugaiska, and Patrick Bonin.
\newblock The stability-plasticity dilemma: Investigating the continuum from
  catastrophic forgetting to age-limited learning effects.
\newblock \emph{Frontiers in psychology}, 4:\penalty0 504, 2013.

\bibitem[Miyato et~al.(2018)Miyato, Maeda, Koyama, and
  Ishii]{miyato2018virtual}
Takeru Miyato, Shin-ichi Maeda, Masanori Koyama, and Shin Ishii.
\newblock Virtual adversarial training: a regularization method for supervised
  and semi-supervised learning.
\newblock \emph{IEEE transactions on pattern analysis and machine
  intelligence}, 41\penalty0 (8):\penalty0 1979--1993, 2018.

\bibitem[Naeini et~al.(2015)Naeini, Cooper, and
  Hauskrecht]{naeini2015obtaining}
Mahdi~Pakdaman Naeini, Gregory Cooper, and Milos Hauskrecht.
\newblock Obtaining well calibrated probabilities using bayesian binning.
\newblock In \emph{Twenty-Ninth AAAI Conference on Artificial Intelligence},
  2015.

\bibitem[Oord et~al.(2018)Oord, Li, and Vinyals]{oord2018representation}
Aaron van~den Oord, Yazhe Li, and Oriol Vinyals.
\newblock Representation learning with contrastive predictive coding.
\newblock \emph{arXiv preprint arXiv:1807.03748}, 2018.

\bibitem[Ovadia et~al.(2019)Ovadia, Fertig, Ren, Nado, Sculley, Nowozin,
  Dillon, Lakshminarayanan, and Snoek]{ovadia2019can}
Yaniv Ovadia, Emily Fertig, Jie Ren, Zachary Nado, David Sculley, Sebastian
  Nowozin, Joshua~V Dillon, Balaji Lakshminarayanan, and Jasper Snoek.
\newblock Can you trust your model's uncertainty? evaluating predictive
  uncertainty under dataset shift.
\newblock \emph{arXiv preprint arXiv:1906.02530}, 2019.

\bibitem[Pan et~al.(2020)Pan, Swaroop, Immer, Eschenhagen, Turner, and
  Khan]{pan2020continual}
Pingbo Pan, Siddharth Swaroop, Alexander Immer, Runa Eschenhagen, Richard
  Turner, and Mohammad Emtiyaz~E Khan.
\newblock Continual deep learning by functional regularisation of memorable
  past.
\newblock \emph{Advances in Neural Information Processing Systems},
  33:\penalty0 4453--4464, 2020.

\bibitem[Parisi et~al.(2019)Parisi, Kemker, Part, Kanan, and
  Wermter]{parisi2019continual}
German~I Parisi, Ronald Kemker, Jose~L Part, Christopher Kanan, and Stefan
  Wermter.
\newblock Continual lifelong learning with neural networks: A review.
\newblock \emph{Neural Networks}, 113:\penalty0 54--71, 2019.

\bibitem[Peng et~al.(2021)Peng, Tang, Jiang, Li, Lei, Lin, and Li]{9354016}
Jian Peng, Bo~Tang, Hao Jiang, Zhuo Li, Yinjie Lei, Tao Lin, and Haifeng Li.
\newblock Overcoming long-term catastrophic forgetting through adversarial
  neural pruning and synaptic consolidation.
\newblock \emph{IEEE Transactions on Neural Networks and Learning Systems},
  pp.\  1--14, 2021.
\newblock \doi{10.1109/TNNLS.2021.3056201}.

\bibitem[Pham et~al.(2021)Pham, Liu, and Hoi]{pham2021dualnet}
Quang Pham, Chenghao Liu, and Steven Hoi.
\newblock Dualnet: Continual learning, fast and slow.
\newblock \emph{Advances in Neural Information Processing Systems}, 34, 2021.

\bibitem[Poole et~al.(2019)Poole, Ozair, Van Den~Oord, Alemi, and
  Tucker]{poole2019variational}
Ben Poole, Sherjil Ozair, Aaron Van Den~Oord, Alex Alemi, and George Tucker.
\newblock On variational bounds of mutual information.
\newblock In \emph{International Conference on Machine Learning}, pp.\
  5171--5180. PMLR, 2019.

\bibitem[Ratcliff(1990)]{ratcliff1990connectionist}
Roger Ratcliff.
\newblock Connectionist models of recognition memory: constraints imposed by
  learning and forgetting functions.
\newblock \emph{Psychological review}, 97\penalty0 (2):\penalty0 285, 1990.

\bibitem[Rebuffi et~al.(2017)Rebuffi, Kolesnikov, Sperl, and Lampert]{icarl}
Sylvestre-Alvise Rebuffi, Alexander Kolesnikov, Georg Sperl, and Christoph~H.
  Lampert.
\newblock icarl: Incremental classifier and representation learning.
\newblock In \emph{Proceedings of the IEEE Conference on Computer Vision and
  Pattern Recognition (CVPR)}, July 2017.

\bibitem[Robins(1995)]{robins1995catastrophic}
Anthony Robins.
\newblock Catastrophic forgetting, rehearsal and pseudorehearsal.
\newblock \emph{Connection Science}, 7\penalty0 (2):\penalty0 123--146, 1995.

\bibitem[Rusu et~al.(2016)Rusu, Rabinowitz, Desjardins, Soyer, Kirkpatrick,
  Kavukcuoglu, Pascanu, and Hadsell]{rusu2016progressive}
Andrei~A Rusu, Neil~C Rabinowitz, Guillaume Desjardins, Hubert Soyer, James
  Kirkpatrick, Koray Kavukcuoglu, Razvan Pascanu, and Raia Hadsell.
\newblock Progressive neural networks.
\newblock \emph{arXiv preprint arXiv:1606.04671}, 2016.

\bibitem[Sajjadi et~al.(2016)Sajjadi, Javanmardi, and
  Tasdizen]{sajjadi2016regularization}
Mehdi Sajjadi, Mehran Javanmardi, and Tolga Tasdizen.
\newblock Regularization with stochastic transformations and perturbations for
  deep semi-supervised learning.
\newblock \emph{Advances in neural information processing systems}, 29, 2016.

\bibitem[Schwarz et~al.(2018)Schwarz, Czarnecki, Luketina, Grabska-Barwinska,
  Teh, Pascanu, and Hadsell]{schwarz2018progress}
Jonathan Schwarz, Wojciech Czarnecki, Jelena Luketina, Agnieszka
  Grabska-Barwinska, Yee~Whye Teh, Razvan Pascanu, and Raia Hadsell.
\newblock Progress \& compress: A scalable framework for continual learning.
\newblock In \emph{International Conference on Machine Learning}, pp.\
  4528--4537. PMLR, 2018.

\bibitem[Titsias et~al.(2020)Titsias, Schwarz, Matthews, Pascanu, and
  Teh]{titsias2020functional}
Michalis~K Titsias, Jonathan Schwarz, Alexander G de~G Matthews, Razvan
  Pascanu, and Yee~Whye Teh.
\newblock Functional regularisation for continual learning with gaussian
  processes.
\newblock In \emph{ICLR}, 2020.

\bibitem[Van~de Ven \& Tolias(2019)Van~de Ven and Tolias]{van2019three}
Gido~M Van~de Ven and Andreas~S Tolias.
\newblock Three scenarios for continual learning.
\newblock \emph{arXiv preprint arXiv:1904.07734}, 2019.

\bibitem[van~de Ven et~al.(2020)van~de Ven, Siegelmann, and
  Tolias]{van2020brain}
Gido~M van~de Ven, Hava~T Siegelmann, and Andreas~S Tolias.
\newblock Brain-inspired replay for continual learning with artificial neural
  networks.
\newblock \emph{Nature communications}, 11\penalty0 (1):\penalty0 1--14, 2020.

\bibitem[Vitter(1985)]{vitter1985random}
Jeffrey~S Vitter.
\newblock Random sampling with a reservoir.
\newblock \emph{ACM Transactions on Mathematical Software (TOMS)}, 11\penalty0
  (1):\penalty0 37--57, 1985.

\bibitem[Xie et~al.(2020)Xie, Dai, Hovy, Luong, and Le]{xie2020unsupervised}
Qizhe Xie, Zihang Dai, Eduard Hovy, Thang Luong, and Quoc Le.
\newblock Unsupervised data augmentation for consistency training.
\newblock \emph{Advances in Neural Information Processing Systems},
  33:\penalty0 6256--6268, 2020.

\bibitem[Zbontar et~al.(2021)Zbontar, Jing, Misra, LeCun, and
  Deny]{zbontar2021barlow}
Jure Zbontar, Li~Jing, Ishan Misra, Yann LeCun, and St{\'e}phane Deny.
\newblock Barlow twins: Self-supervised learning via redundancy reduction.
\newblock \emph{arXiv preprint arXiv:2103.03230}, 2021.

\bibitem[Zenke et~al.(2017)Zenke, Poole, and Ganguli]{zenke2017continual}
Friedemann Zenke, Ben Poole, and Surya Ganguli.
\newblock Continual learning through synaptic intelligence.
\newblock In \emph{International Conference on Machine Learning}, pp.\
  3987--3995. PMLR, 2017.

\bibitem[Zhai et~al.(2019)Zhai, Oliver, Kolesnikov, and Beyer]{zhai2019s4l}
Xiaohua Zhai, Avital Oliver, Alexander Kolesnikov, and Lucas Beyer.
\newblock S4l: Self-supervised semi-supervised learning.
\newblock In \emph{Proceedings of the IEEE/CVF International Conference on
  Computer Vision}, pp.\  1476--1485, 2019.

\end{thebibliography}
\bibliographystyle{collas2022_conference}

% \newpage
\appendix

\section{Implementation details}

\subsection{Datasets} \label{datasets}
Following \citep{buzzega2020dark}, we evaluate on the following CL scenarios:
 
\textbf{Class-IL}: The CL model encounters a new set of classes in each task and must learn to distinguish all classes encountered thus far after each task. In practice, we split CIFAR-10 \citep{cifar10}, CIFAR-100 \citep{cifar10}, and TinyImageNet \citep{tinyimagenet} into partitions of 2, 20, and 20 classes per task, respectively.  

% \textbf{Task Incremental Learning (Task-IL)}, although similar to Class-IL, accesses task identities to select relevant classifier for each data sample. Task-IL is not truly representative of the real world scenario. 

\textbf{Domain-IL}: The number of classes remain the same across subsequent tasks. However, a task-dependent transformation is applied changing the input distribution for each task. Specifically, R-MNIST \citep{gem} rotates the input images by a random angle in the interval $[0; \pi)$. R-MNIST requires the model to classify all 10 MNIST \citep{mnist} digits for 20 subsequent tasks. Permuted-MNIST \citep{kirkpatrick2017overcoming}, on the other hand, applies random permutation to the pixels. Both Rotated and Permuted MNIST require the model to classify all 20 MNIST \citep{mnist} digits for 20 subsequent tasks. 
Permuted-MNIST violates cross-task resemblance desiderata, therefore, not considered in our work. 

\textbf{MNIST-360} \citep{buzzega2020dark} models a stream of MNIST data with batches of two consecutive digits at a time. Each sample is rotated by an increasing angle and the sequence is repeated six times. MNIST-360 exposes the CL model to both sharp class distribution shift and smooth rotational distribution shift.

\subsection{Derivation of regularizers}\label{app_ssl_algo}
In this section, we provide implementation details of different consistency regularizers. 

\subsubsection{$L_p$ family}
Derivation of ${L}_p$ family of loss functions is straight forward. We consider ${L}_1$, ${L}_2$ and ${L}_{inf}$ in our comparisons. These variants can be easily obtained by replacing $p$ in Equation \ref{eqn_lp}.

\subsubsection{KL-Divergence}
KL-Divergence is a non-symmetric measure of the difference between two probability distributions $P(Z_\theta)$ and $Q(Z_r)$, is closely related to relative entropy, information divergence, and information for discrimination. In order to obtain $P$ and $Q$, we apply softmax operation on the predictions.

\subsubsection{Self-supervised learning based regularizers}
A broad categorization of leading SSL methods can be found in Table \ref{tab:ssl_algorithms}. For all these methods, predictions are $L_2$ normalized before being fed into the respective loss function.  We employ one algorithm from each category in our comparison. 

We employ SimCLR \citep{chen2020simple} from InfoNCE as in Equation \ref{eqn_infonce}.  Several approaches have been proposed to reduce the reliance on large negatives in contrastive learning. BYOL \citep{grill2020bootstrap} and SimSiam \citep{chen2021exploring} addressed this lacuna through negative-free cosine similarity loss. In our implementation, BYOL can be instantiated as follow:
\begin{equation}
\label{eqn_byol}
     \mathcal{L}_{sc} =  2 - 2 \frac{<z_\theta, z_r>}{\lVert z_\theta \rVert_2 \; \lVert z_r \rVert_2}
\end{equation}

On the other hand, clustering-based method DINO \citep{Caron_2021_ICCV}) creates a proxy task through cross-entropy loss. To keep our design simple, we do not use centering operation through momentum update. We simply treat stored predictions as target by applying softmax operation on both predictions. 
Barlow-Twins \citep{zbontar2021barlow} proposed an objective function measuring the cross-correlation matrix  and making it as close to the identity matrix as possible. The objective function for Barlow Twins can be defined as:
\begin{equation}
\label{eqn_barlow_twins}
    \mathcal{L}_{sc} = \sum_i (1 - C_{ii})^2 + \sum_i \sum_{i\ne j} C_{ij}^2
\end{equation}
where $C$ is the cross-correlation matrix computed between the predictions along the batch dimension. 
 
Despite the technical variety, most if not all SSL methods are prone to trivial, collapsed solutions by design \citep{hua2021feature}. Learning leads to degenerative solutions when all the representations collapse to a single point in the representation space. Our framework inherently avoids degenerative solutions: cross-entropy objective makes sure that representations do not collapse to a single point. Therefore, stored representations are scattered across the representation space. Since our framework tries to improve affinity between stored representations and representations generated through current model, all positive pairs cannot collapse to a single point, thereby avoiding total collapse without the helps of ad-hoc techniques.

\subsection{Relation to contemporary methods}\label{byol}
In this section, we review several works that are closely related to our proposal in CL.

\textbf{Knowledge distillation:} Several approaches employ past version of the CL model as a teacher and distill knowledge to the current model. Learning without forgetting (LwF) \citep{lwf} reduces the feature drift by smoothening the predictions at the beginning of each task without employing the memory buffer. On the other hand, iCaRL \citep{icarl} combines replay with knowledge distillation. iCaRL trains a nearest-mean-of-exemplars classifier using memory buffer as a training set and mitigates catastrophic forgetting through knowledge distillation. In our proposal however, we do not use any previous checkpoints as a teacher. As storing predictions alongside input images has a limited memory footprint, we opt for storing and replaying logits to ensure consistency in predictions.

\textbf{Function regularization:} Similar to our proposal, several methods (e.g. \citet{titsias2020functional, pan2020continual, fdr, buzzega2020dark}) regularize the change in the function learned by the CL model. \citet{titsias2020functional} constrains the neural network predictions from deviating too far from those that solve previous tasks  through KL-Divergence. Function Distance Regularization (FDR) \citep{fdr} uses past exemplars and model outputs to align past and current outputs. However, FDR stores model outputs at task boundaries. Dark-Experience Replay (DER++) \citep{buzzega2020dark} extended the work in FDR by introducing a \textit{reservoir} thereby eliminating the need for task boundaries without experiencing a drop in performance. Although these approaches use some form of consistency regularization, their effectiveness has not been verified under a common framework. 

% In this section, we show how our proposed framework subsumes DER++ under special circumstances. In addition to cross-entropy losses, DER++ employs squared error ($\mathcal{L}_{mse}=\lVert z_\theta - z_r \rVert_2^2$) to align past and current outputs. When juxtaposed against BYOL in our framework, the main difference between  maximizing cosine similarity ($\mathcal{L}_{cos} = \frac{z_\theta^T z_r}{\lVert z_\theta\rVert_2\lVert z_r\rVert_2}$) and minimizing squared error is whether or not the vectors are normalized to unit L2-norms i.e.
% \begin{equation}
%     \mathcal{L}_{se} =  2 - 2 \; \mathcal{L}_{cos} 
% \end{equation}
% where $\lVert z_\theta\rVert_2 = 1$ and $ \lVert z_r\rVert_2 = 1$. Therefore, maximizing cosine similarity between past and current outputs using BYOL in our framework translates to minimizing squared error between them in DER++. 

\textbf{Latent replay:} Consistency regularization in rehearsal-based CL can also be connected to generative-rehearsal based methods \citep{van2020brain, caccia2020online} with latent representation replay. As an alternative to storing data in the buffer, these methods focus on generating the data to be replayed with a learned generative neural network model of past observations. In addition, they leverage temperature-scaled soft targets for the replayed samples to distill the knowledge. In our case, we use past predictions themselves as a soft targets and enforce the current predictions to be aligned with them. 

\textbf{Intersection of SSL and CL:} There have been several works \citep{madaan2021representational, fini2021self, davari2022probing} that fall in the intersection of SSL and CL. \citet{fini2021self} showed that self-supervised loss functions can be seamlessly converted into distillation losses for CL by adding a predictor network that maps the current state of the representations to their past state. However, this method requires that the past state of the CL model be stored at the end of each task. \citet{pham2021dualnet} employed SSL for consolidating generalizable features in a slow-learning network in a complementary learning system inspired CL setup. Our proposal drastically differs from these approaches: We do not use SSL as a primary learning objective. Instead, SSL objective functions are used to enforce consistency in predictions across time.

\section{Task performance}
\begin{figure*}[!hbt]
\centering
  \includegraphics[width=1\linewidth]{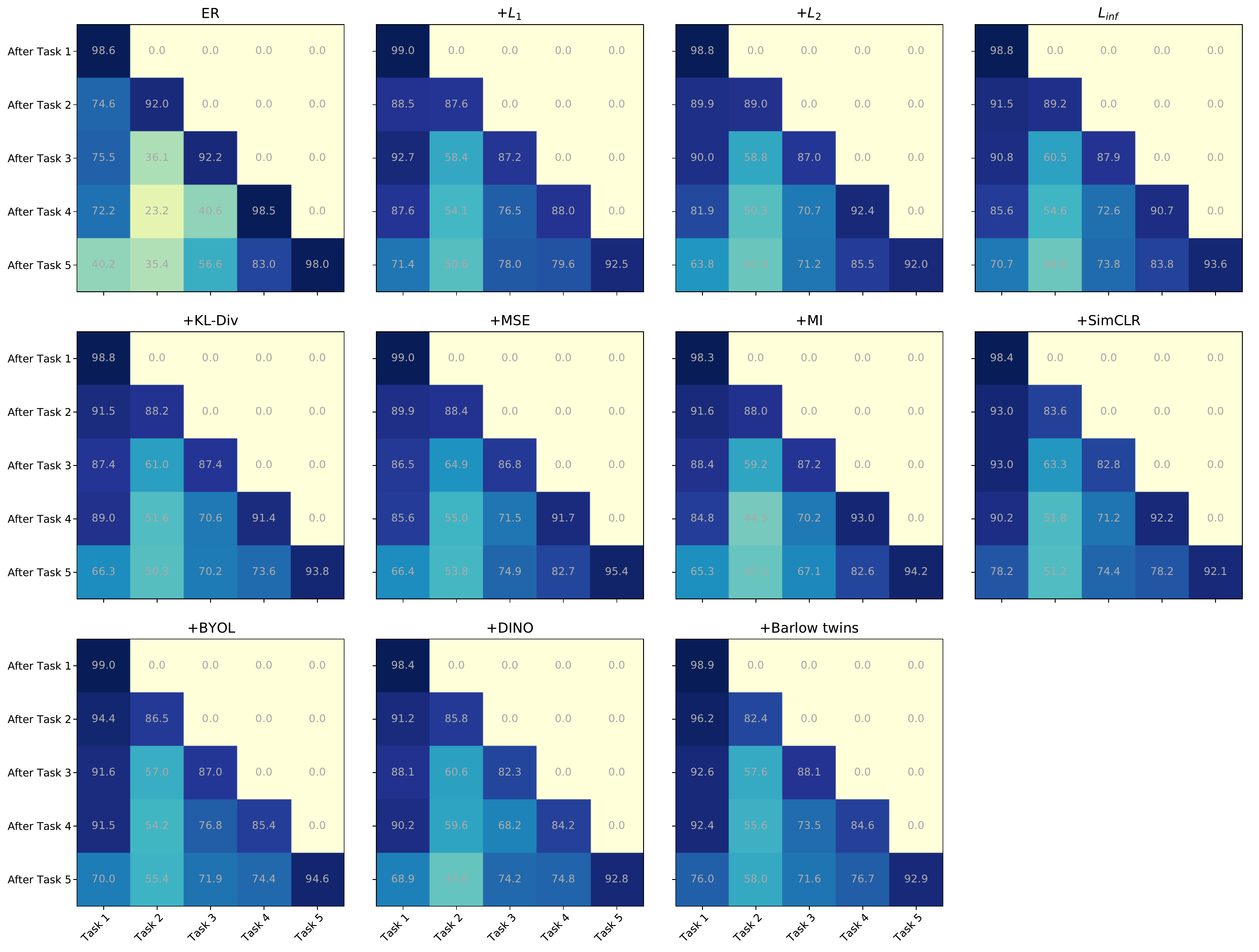}
  \caption{Task performance of different CL models on S-CIFAR-10 with buffer size 500. }
  \label{fig:tasks_acc_appendix}
\end{figure*}

 We provide detailed task performance throughout the CL training in Figure \ref{fig:tasks_acc_appendix}. The CL models are trained on the S-CIFAR10 with the buffer size 500. These results supplement the results presented in Table \ref{tab:main} where only the average of accuracies after last task are presented. As can be seen, Regularizing consistency in predictions consistently augments ER throughout the CL training. 

\section{Model Analysis on S-TinyImageNet}
We extend the CL model analysis to S-TinyImageNet in which the models are trained on this data with buffer size 500. 

\subsection{Model Calibration}
\begin{figure*}[!hbt]
\centering
  \includegraphics[width=1\linewidth]{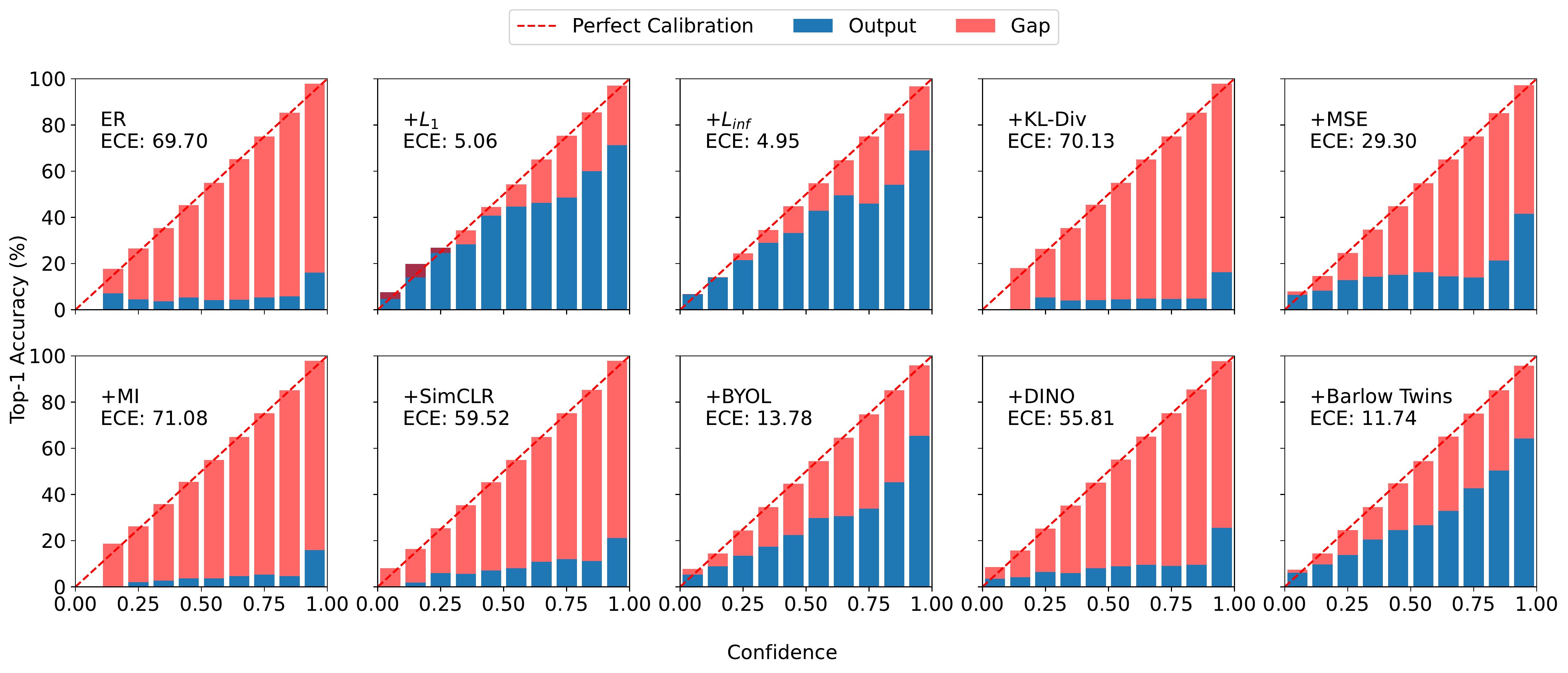}
  \caption{Confidence estimates and corresponding Expected Calibration Error (ECE) of S-TinyImageNet trained CL models. Lower ECE is better.}
  \label{fig:calib_tiny}
\end{figure*}

Figure \ref{fig:calib_tiny} presents the reliability diagrams of CL models along with corresponding ECE score. As is the case in Section \ref{sec:calib}, ER is highly miscalibrated. Owing to the increased complexity in S-TinyImageNet, the miscalibration is much more pronounced in ER. Except MI, all other consistency regularizers improve model calibration. Especially, $L_{1}$ and $L_{inf}$ greatly improve the model calibration in S-TinyImageNet, reinforcing our earlier finding that stricter consistency constraints assist ER better than approximate counterparts. 

\subsection{Task-recency bias}
\begin{figure*}[!hbt]
\centering
  \includegraphics[width=1\linewidth]{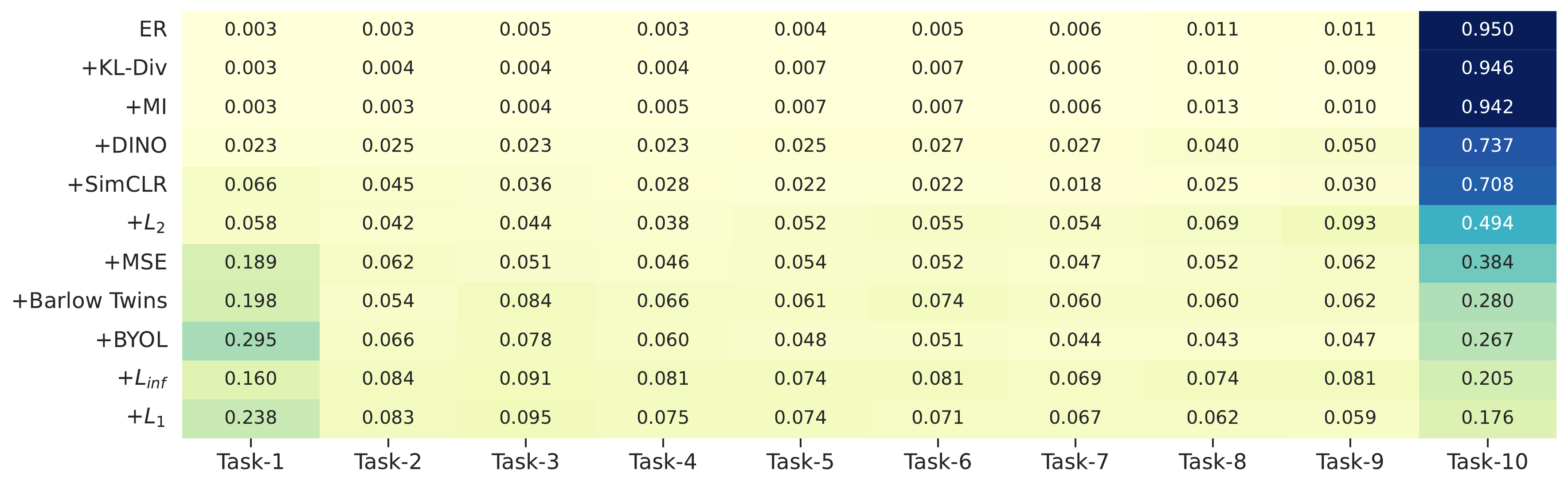}
  \caption{Average task probabilities of CL models trained on S-TinyImageNet with 500 buffer size. Methods are sorted in descending order of their bias towards the most recent task. ER has the highest task recency bias while stricter consistency constraints have one of the lowest biases.}
  \label{fig:task_prob_tiny}
\end{figure*}

 Figure \ref{fig:task_prob_tiny} presents the the normalized probabilities of each task of a S-CIFAR-10 trained model, computed by averaging probabilities of all samples belonging to the associated classes in a Class-IL setting.   The predictions in ER are biased mostly towards recent tasks, with most recent task being almost $0.95$.  Therefore, the predictions stored in the buffer are completely biased towards their corresponding task logits. Since consistency regularization penalizes any deviation in predictions, recency bias is mitigated as a by-product. Stricter consistency constraints such as +$L_1$, +$L_{inf}$ have a lower bias. Among SSL algorithms, +Barlow Twins and +BYOL have lower task recency bias.
 
\subsection{Overview of S-TinyImageNet evaluation}\label{appendix_overview}
We provide an overview of relative performance gain of different regularizers  trained on S-TinyImageNet with buffer size 500 over ER in Figure \ref{fig:overview}. We provide relative performance gain in terms of accuracy, task-recency bias and model calibration. The relative gains are computed as follows:
\begin{equation}
    Accuracy \; Gain = 100 * (Acc_{cr} - Acc_{er}) / Acc_{er}
\end{equation}
\begin{equation}
    Recency \; bias \; Gain = 100 * ((1- (T10_{cr} - T1_{cr})) - (1- (T10_{er} - T1_{er}))) / (1- (T10_{er} - T1_{er}))
\end{equation}
\begin{equation}
    Calibration \; Gain = 100 * ((100-ECE_{cr}) -(100-ECE_{er})) / (100-ECE_{er})
\end{equation}
 where $cr$ refers to performance of any consistency regularizer, $T10$ and $T1$ are average task probabilities of task 10 and task 1.

\section{Forward Transfer on R-MNIST}
 We report the forward transfer results on R-MNIST experiments presented earlier in Table \ref{tab:main}. Following \citep{gem}, we compute forward transfer as a difference between the accuracy just before starting training on a given task and the one of the random-initialized network; it is averaged across all tasks. Forward transfer is relevant for Domain-IL scenarios as long as underlying input transformation is not disruptive. Table \ref{tab:fwt} shows the forward transfer on R-MNIST for different buffer sizes. Higher forward transfer is a desirable property as information consolidated in the previous tasks can be leveraged for learning new tasks. All methods including vanilla ER have positive forward transfer in R-MNIST. Both +MSE and +$L_{inf}$ have higher positive transfer while approximate methods such as +MI and +SimCLR have one of the lowest forward transfers. Strict but simple consistency regularizers improve positive forward transfer across buffer sizes thereby boosting the overall performance.

 \begin{table}[H]
 \center
\caption{Forward transfer on R-MNIST}
\label{tab:fwt}

    \begin{tabular}{l|c|c}
    \toprule
    Method & Buffer size = 200 & Buffer size = 500  \\ 
    \midrule
    ER	& 60.30\scriptsize{$\pm$1.32} & 63.82 \scriptsize{$\pm$2.50} \\
    +$L_{1}$ & 59.96   \scriptsize{$\pm$1.23}  & 64.67   \scriptsize{$\pm$0.69}\\
     +$L_{2}$ & 63.68 \scriptsize{$\pm$0.93} &  66.80 \scriptsize{$\pm$0.60}\\
     +$L_{inf}$ & 	\underline{64.08} \scriptsize{$\pm$1.03} &   \underline{67.30} \scriptsize{$\pm$0.14} \\
     +KL-Div & 	58.29 \scriptsize{$\pm$2.71} &  62.84 \scriptsize{$\pm$0.53}\\
     +MSE & \textbf{64.87} \scriptsize{$\pm$1.02} &    	\textbf{67.42} \scriptsize{$\pm$0.43} \\
    +MI	& 60.05 \scriptsize{$\pm$2.15} & 63.61 \scriptsize{$\pm$0.32}	\\
    +SimCLR	& 59.44 \scriptsize{$\pm$2.29} & 63.16 \scriptsize{$\pm$0.23}	\\
    +BYOL	& 63.04 \scriptsize{$\pm$1.39} & 65.86  \scriptsize{$\pm$0.71} \\
    +DINO	& 60.77   \scriptsize{$\pm$1.98} & 62.13 \scriptsize{$\pm$1.79} 	\\
    +Barlow Twins	& 62.46  \scriptsize{$\pm$1.51} & 65.63 \scriptsize{$\pm$0.08}  \\
    \bottomrule
    \end{tabular}
\end{table}

\section{Expected calibration error}\label{ece_computation}
Miscalibration in neural networks can be defined as a  difference in expectation between confidence and accuracy \citep{guo2017calibration} i.e.
\begin{equation}
\label{ece_eqn}
\underset{\hat{P}}{\mathbb{E}}[|\mathbb{P}(\hat{Y}=Y \mid \hat{P}=p)-p|]
\end{equation}
where $\hat{Y}$ is a class prediction and $\hat{P}$ is its associated confidence, i.e. probability of correctness.  ECE approximates Equation \ref{ece_eqn} by partitioning the predictions into M equal spaced bins and taking a weighted average of the bins’ accuracy/confidence
difference i.e.
\begin{equation}
\mathrm{ECE}=\sum_{m=1}^{M} \frac{\left|B_{m}\right|}{n}\left|\operatorname{acc}\left(B_{m}\right)-\operatorname{conf}\left(B_{m}\right)\right|
\end{equation}
where $n$ is the number of samples. Lower ECE indicates better calibration of the model.

\section{Effect of $\alpha$ on the performance of ER}
We test the robustness of our analysis by tuning  $\alpha$ in our experiments. One might also wonder that  a simple fix to the task-recency bias in ER is to increase the magnitude of  $\alpha$. Table \ref{tab:alpha} presents Effect of $\alpha$ on vanilla ER's performance on S-CIFAR10 with buffer size 500. One can see that increasing the magnitude of the  $\alpha$ disproportionately affects the overall performance (i.e., accuracy, calibration). Therefore, increasing the magnitude of $\alpha$ does not yield any discernible benefits including reduction in task recency bias. 
 
\begin{table}[H]
 \center
\caption{Effect of $\alpha$ on vanilla ER's performance on S-CIFAR10 with buffer size 500.}
\label{tab:alpha}

    \begin{tabular}{c|c|c|c}
    \toprule
     $\alpha$ & Accuracy Top-1 ($\%$) & ECE score & Task-5 probability \\
    \toprule
    1 	&  \textbf{62.65} &  \underline{31.8} & \textbf{0.456} \\
    2 	 & 55.14 & 33.73 & 0.5291 \\
    5 	&  \underline{56.14}  & \textbf{31.74} &  \underline{0.458} \\
    10 	& 47.62 & 39.67 & 0.489  \\
    \bottomrule
    \end{tabular}
\end{table}

\section{Sensitivity of approximate methods to $\beta$}
Our analysis shows that strict but simple regularizers perform better than approximate methods considered in this work. However, one might wonder whether the approximate methods perform better if their penalty ($\beta$) were to be increased. Table \ref{tab:beta} shows the effect of $\beta$ on the performance of approximate methods on TinyImageNet with buffer size 500. As can be seen, the performance of approximate methods is still quite inferior to the strict consistency constraints. The results presented in our work therefore are robust to the choice of $\beta$.

\begin{table}[H]
 \center
\caption{Effect of $\beta$ on the performance of approximate methods on TinyImageNet with buffer size 500.}
\label{tab:beta}

    \begin{tabular}{c|c|c|c}
    \toprule
     $\beta$ & +MI & +SimCLR & +DINO \\
    \toprule
    0.05 & 9.93 \scriptsize{$\pm$0.21} &  13.05 \scriptsize{$\pm$0.08} & 13.04 \scriptsize{$\pm$0.84 } \\
    0.1 & 10.08 \scriptsize{$\pm$0.28}  & 13.93 \scriptsize{$\pm$0.84} & 13.91 \scriptsize{$\pm$0.55}  \\
    0.5 & 10.26 \scriptsize{$\pm$0.26}&  14.81 \scriptsize{$\pm$0.79}&  14.23 \scriptsize{$\pm$0.54}  \\
    1 & 10.41 \scriptsize{$\pm$0.40}&  14.68 \scriptsize{$\pm$0.81}&  14.12 \scriptsize{$\pm$0.46}  \\
    1.5 & 10.33 \scriptsize{$\pm$0.37}&  14.57 \scriptsize{$\pm$0.50}&  13.55 \scriptsize{$\pm$0.56} \\
    2 & 10.56 \scriptsize{$\pm$0.28}&  14.00 \scriptsize{$\pm$0.63}&  13.81 \scriptsize{$\pm$0.81}  \\
    5 & 10.32 \scriptsize{$\pm$0.13}&  13.15 \scriptsize{$\pm$0.74}&  11.44 \scriptsize{$\pm$0.92}  \\
    10 & 9.84 \scriptsize{$\pm$0.33}&  11.70 \scriptsize{$\pm$0.82}&  9.60 \scriptsize{$\pm$0.49} \\
    50 & 7.78 \scriptsize{$\pm$0.36}&  7.96 \scriptsize{$\pm$0.41}&  8.62 \scriptsize{$\pm$0.25}  \\
    \bottomrule
    \end{tabular}
\end{table}

\end{document}